%% file: root.tex
\documentclass[letterpaper, 10 pt, conference]{ieeeconf}  

\IEEEoverridecommandlockouts                              

\input{preamble}

\title{\LARGE \bf
Exploration-assisted Bottleneck Transition Toward Robust and Data-efficient Deformable Object Manipulation
}

\author{Yujiro Onishi$^{1}$, Ryo Takizawa$^{1}$, Yoshiyuki Ohmura$^{1}$, Yasuo Kuniyoshi$^{1}$
\thanks{The authors are with the Mechano-Informatics Department, Graduate School of Information Science and Technology, The University of Tokyo, Tokyo 113-8656, Japan.
(e-mail: y-onishi@isi.imi.i.u-tokyo.ac.jp; takizawa@isi.imi.i.u-tokyo.ac.jp; ohmura@isi.imi.i.u-tokyo.ac.jp; kuniyosh@isi.imi.i.u-tokyo.ac.jp).}
}

\makeatletter
\let\@oldmaketitle\@maketitle 
\renewcommand{\@maketitle}{%
  \@oldmaketitle 
  \setcounter{figure}{0} 
  \vspace{2mm} 
  \centering
  \includegraphics[width=0.93\linewidth]{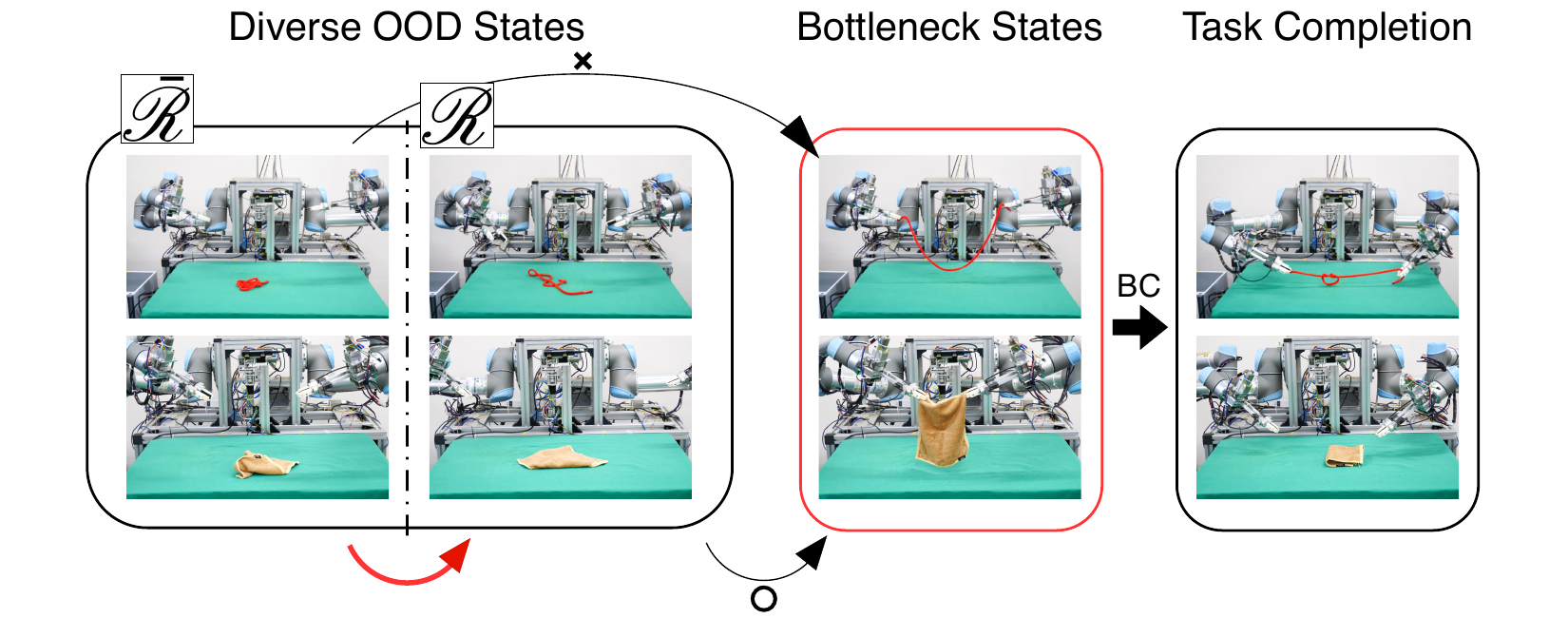}
  \captionof{figure}{%
    \textbf{Exploration-assisted Bottleneck Transition for Deformable Object Manipulation (ExBot).}
    ExBot enables robust rope and cloth manipulation by introducing exploration-guided transitions through bottleneck states.
    Our framework successfully completes learned tasks from the out-of-distribution (OOD) states, including severe self-occluded or crumpled conditions. 
    The red arrow represents exploration action, and the red frame represents bottleneck states. This mechanism promotes both data-efficiency and robustness in deformable object manipulation.
  }
  \label{fig:teaser}
  \vspace{0mm}
}
\makeatother

\begin{document}
\maketitle
\thispagestyle{empty}
\pagestyle{empty}

\begin{abstract}
Imitation learning has demonstrated impressive results in robotic manipulation but fails under out-of-distribution (OOD) states. This limitation is particularly critical in Deformable Object Manipulation (DOM), where the near-infinite possible configurations render comprehensive data collection infeasible. Although several methods address OOD states, they typically require exhaustive data or highly precise perception. Such requirements are often impractical for DOM owing to its inherent complexities, including self-occlusion.
To address the OOD problem in DOM, we propose a novel framework, Exploration-assisted Bottleneck Transition for Deformable Object Manipulation (ExBot), which addresses the OOD challenge through two key advantages. First, we introduce bottleneck states, standardized configurations that serve as starting points for task execution. This enables the reconceptualization of OOD challenges as the problem of transitioning diverse initial states to these bottleneck states, significantly reducing demonstration requirements. Second, to account for imperfect perception, we partition the OOD state space based on recognizability and employ dual action primitives. This approach enables ExBot to manipulate even unrecognizable states without requiring accurate perception. By concentrating demonstrations around bottleneck states and leveraging exploration to alter perceptual conditions, ExBot achieves both data efficiency and robustness to severe OOD scenarios.
Real-world experiments on rope and cloth manipulation demonstrate successful task completion from diverse OOD states, including severe self-occlusions.
\end{abstract}

\section{INTRODUCTION}
Recent advances in imitation learning (IL) have demonstrated remarkable success in enabling robots to perform complex dexterous tasks by learning policies directly from human demonstrations \cite{zhao2023learning, Kim2024GC-DA, Zhao2025AlohaUnleashed}. However, IL methods are inherently vulnerable to distribution shift \cite{osa2018algorithmic}. This challenge is particularly acute in deformable object manipulation (DOM), where the near-infinite variety of possible configurations renders it practically infeasible to collect exhaustive training data or to rely on precise analytical models.

To bridge the gap between the efficiency of IL and the necessity for robustness in out-of-distribution (OOD) scenarios, we propose a novel framework: Exploration-assisted Bottleneck Transition for Deformable Object Manipulation (\textbf{ExBot}). The core of our approach lies in the introduction of ``bottleneck states'', standardized configurations that serve as reliable starting points for task execution. By establishing these bottleneck states, we reformulate the OOD challenge into a transition task, where arbitrary initial states are guided toward a known bottleneck. This formulation effectively shifts the OOD challenge from learning to generalize across all states to a perceptual problem: identifying the current configuration and taking the appropriate transition action.

Existing perception algorithms for DOM frequently fail in the 
presence of severe self-occlusion and inherent visual 
ambiguity. Perfect perception is often unattainable in DOM; 
hence, ExBot departs from the pursuit of state estimation and 
instead \textit{embraces} imperfect perception by partitioning the OOD state space based on recognizability. For sufficiently recognizable states, the robot executes ``Preparation Action'' to transition the object toward a bottleneck state. Conversely, for unrecognizable states where perception fails, the system initiates ``Exploration Action'', dynamic physical interventions designed to drastically alter the object's configuration until it becomes recognizable. To assess recognizability, we leverage vision-language models (VLMs). This approach enables ExBot to achieve robust task completion without requiring exhaustive modeling or prohibitive amounts of training data.

\section{RELATED WORK}
\subsection{Imitation Learning and Out-of-Distribution Challenges}
Recent research in Imitation Learning (IL) has demonstrated impressive capabilities in acquiring new skills from human demonstrations \cite{zhao2023learning, Kim2024GC-DA, Zhao2025AlohaUnleashed, chi2025diffusion}.
However, visuomotor policies trained with IL often exhibit out-of-distribution (OOD) vulnerability \cite{osa2018algorithmic}. Several approaches have been proposed to address the OOD problem. Collecting enormous amounts of diverse demonstration data \cite{lin2024data} is particularly unsuitable for DOM due to the near-infinite variety of configurations, rendering exhaustive data collection infeasible. Leveraging relative coordinates from gaze inference facilitates manipulation that is robust to OOD shifts \cite{takizawa2025enhancing}. This robustness applies to changes in a rigid object's position and orientation. However, this approach faces limitations when applied to diverse states of deformable objects involving self-occlusion owing to the inherent diversity of bottleneck states and the challenges posed by unobservable target points. Learning dynamics models using simulators \cite{Ruihai2025NeuralDA}, or learning recovery policies that transition from OOD to in-distribution states \cite{gao2024out} are closely related to our insight of guiding the system back to the training data region when encountering OOD states. However, these approaches are infeasible for DOM due to the difficulty of obtaining precise object models.
Furthermore, these methods assume reliable perception that is frequently violated in DOM, where perception often fails due to severe occlusions and complex deformations.

Instead, this study reconceptualizes the OOD problem by introducing bottleneck states, a subset of the entire task state space that serves as standardized starting points for task execution.
This concept offers two key advantages for DOM: first, data-efficient learning by requiring demonstrations only from bottleneck states rather than covering the entire state space. Second, a problem decomposition that separates the challenge into transitioning to bottleneck states and executing tasks from these standardized configurations.
The results show that bottleneck states can achieve both data efficiency and robustness for DOM.

\subsection{Perception for Deformable Objects}
Perception of deformable objects remains challenging due to the high-dimensional state space and complex deformations \cite{yin2021modeling}. Numerous research have focused on improving state estimation algorithms \cite{xiang2023trackdlo, choi2024mBEST, lips2024learning}. However, these methods struggle with configurations that inherently resist accurate perception, such as crumpled cloth or tangled ropes.
Considering this challenge, our work takes a different stance. In this study, rather than pursuing perfect perception, our framework considers imperfect perception as a fundamental principle. When the system determines that the current state is unrecognizable, we employ Exploration Actions that physically alter object configurations drastically to reach recognizable states.
A closely related work is MANIP~\cite{yu2024manip}, which also utilizes actions to improve perception. However, our approach differs in two key aspects. 
First, we explicitly partition the OOD state space into recognizable and unrecognizable states, whereas MANIP characterizes states with a continuous confidence vector without making this binary distinction. Second, MANIP’s uncertainty quantification relies on heatmaps produced by a learned model, indicating that its effectiveness is contingent on the underlying model operating within its training distribution. Consequently, MANIP’s interactive perception policies can only meaningfully reduce uncertainty where the learned model still produces valid heatmap outputs. In states where the recognition algorithm fundamentally fails (e.g., severe self-occlusion, crumpling), the heatmap becomes uninformative, rendering the uncertainty metric unreliable as a recovery signal. In contrast, our Exploration Actions explicitly target such unrecognizable states by physically restructuring the object until the recognition algorithm’s operative assumptions are restored.

\section{METHOD}

\subsection{Problem Formulation}
Our objective is to enable robust task execution from diverse out-of-distribution initial states of deformable objects.

Let $\mathcal{S}$ denote the entire state space of possible object configurations. 
For a given manipulation task, we define \textbf{bottleneck states} $\mathcal{S}_b \subset \mathcal{S}$ as a pre-determined subset of configurations characterized by low variance. This is designed empirically and established before data collection.
Moreover, this corresponds to a focused region of the state space from which a learned policy can reliably execute the task with high success rates. 
Practically, bottleneck states serve as starting points for data collection, providing a standardized set of initial configurations that ensure consistent task performance across multiple trials.

Crucially, transitioning to a bottleneck state not only standardizes the object configuration but also brings the robot's joint configuration into a range covered by the training data distribution. Therefore, the downstream policy can operate in a familiar state space for both the object and the robot, contributing to the data efficiency of our approach.

Learning manipulation from all possible initial states requires prohibitive amounts of training data for high-dimensional deformable objects. Instead, we decompose the problem into two stages:
\begin{enumerate}
    \item \textbf{Transition to a bottleneck state}: Transform arbitrary initial states $s \in \mathcal{S}$ into a configuration $s_b \in \mathcal{S}_b$.
    \item \textbf{Task execution from a bottleneck state}: Execute the manipulation task from $s_b$ using a learned policy
\end{enumerate}
This decomposition enables data-efficient learning: demonstrations only need to cover the trajectory from $s_b$ to task completion. The first stage handles the diversity of initial configurations through modular transition primitives. This is particularly valuable for deformable objects, where the state space is intractably large and comprehensive data collection is practically impossible.

\begin{figure}[t]
    \centering
    \includegraphics[width=0.9\columnwidth]{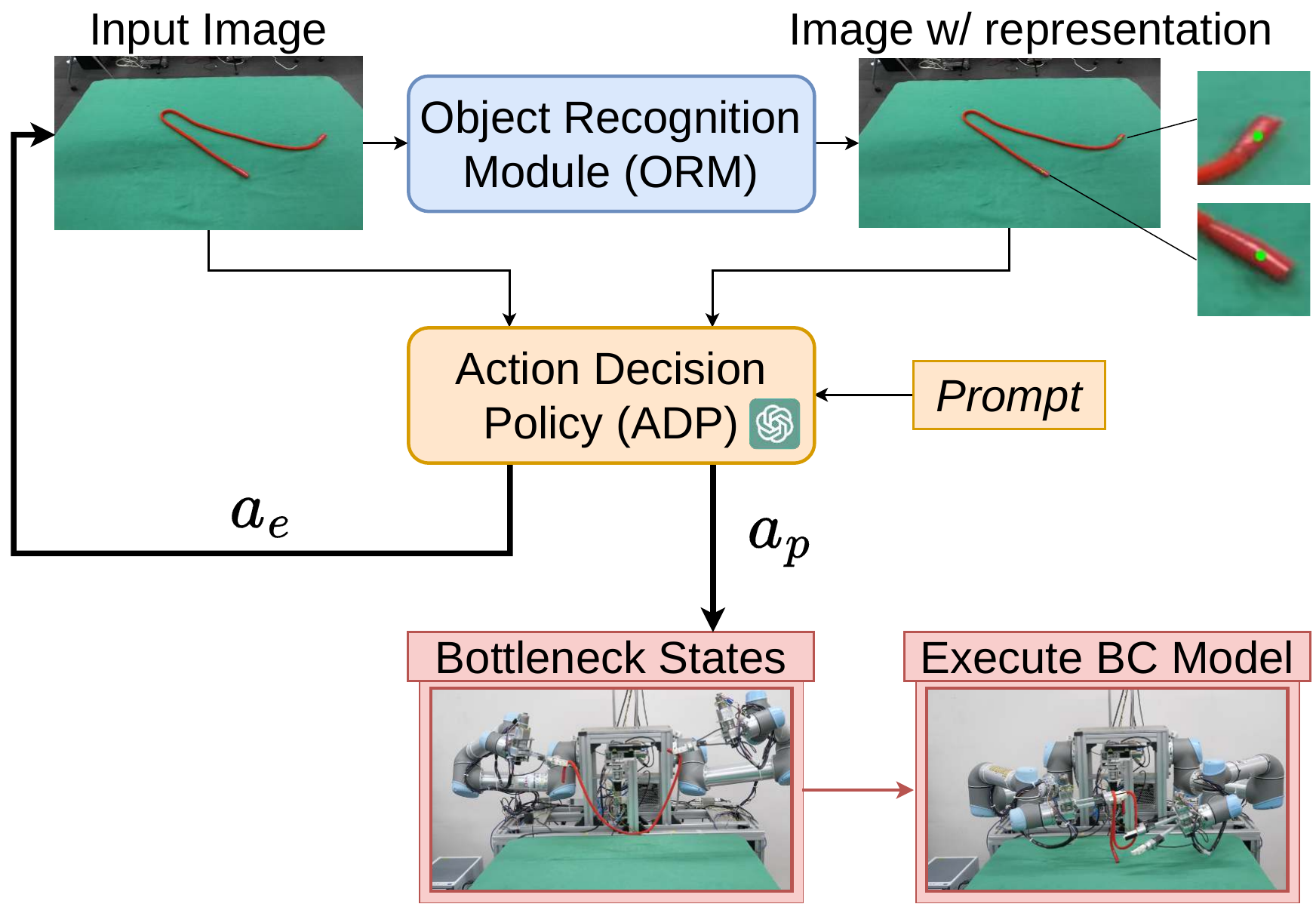}
    \caption{System architecture. The Object Recognition Module and Action Decision Policy handle OOD states by selecting between Preparation Action $a_p$ and Exploration Action $a_e$, transitioning the object into a bottleneck state. Once t his transition is conducted, the Behavior Cloning (BC) Model executes the tasks from the bottleneck state.}
    \label{fig:architecture}
    \vspace{-0.5em}
\end{figure}

\subsection{State Classification Based on Recognizability}
\label{subsec:def_recognizability}

To realize transitions to bottleneck states while avoiding learning from diverse OOD states, we design the system to extract abstract object keypoint representations and plan appropriate actions based on these representations. However, recognition algorithms often fail to extract accurate representations due to severe self-occlusion. To achieve robust deformable object manipulation, the system must be capable of handling states beyond successful recognition.

To handle these states, we partition the state space based on the \textit{recognizability} of object representations by the recognition algorithm:

\begin{itemize}
    \item \textbf{Recognizable states} $\mathcal{R}$:  States where the recognition algorithm successfully extracts the intended object representations
    \item \textbf{Unrecognizable states} $\bar{\mathcal{R}}$: States where extraction fails or produces incorrect representations due to algorithmic limitations or assumptions
\end{itemize}

Based on this partition, we design two types of action primitives:
\begin{align}
\text{Preparation Action } a_p &: \mathcal{R} \rightarrow \mathcal{S}_b \\
\text{Exploration Action } a_e &: \bar{\mathcal{R}} \rightarrow \mathcal{R}
\end{align}

\begin{figure}[t]
\centering
    \includegraphics[width=0.85\linewidth]{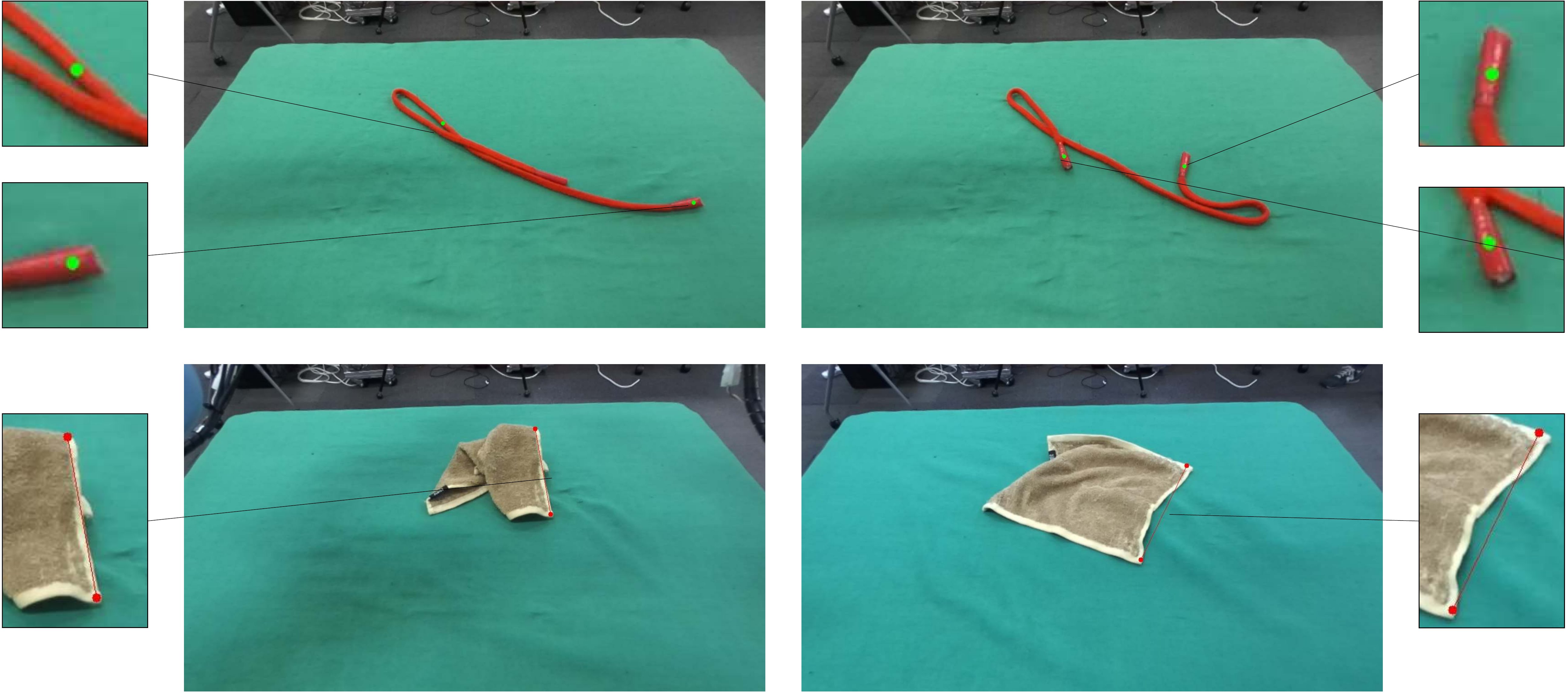}
    \caption{Examples of images with keypoint representations for input of the Action Decision Policy, which receives both the raw image and the image with representation. (Top row) Rope manipulation: two endpoints are extracted. (Bottom row) Cloth manipulation: represents two adjacent corners with an adjacent guideline are represented.}
    \label{fig:recog}
\end{figure}

\begin{figure}[t]
\begin{promptbox}
    \footnotesize
    \textbf{Task:} [Verification objective]\\[0.5em]
    \textbf{Input data:} [Explanations of each input image]\\[0.5em]
    \textbf{Conditions for correct representation:}
    \begin{itemize}[leftmargin=1.5em, itemsep=0pt, topsep=2pt, parsep=0pt]
        \item Condition 1: ...
        \item Condition 2: ...
    \end{itemize}
    \textbf{Output format:}\\
    reasoning: \textcolor{gray}{Provide bullet-point analysis}\\
    answer: \textcolor{gray}{Output either ``ANSWER: YES'' or ``ANSWER: NO''}
\end{promptbox}
\caption{General VLM prompt structure for verification tasks.}
\label{fig:vlm_example}
\vspace{-0.7em}
\end{figure}

\textbf{Preparation Action} ($a_p$) transitions recognizable states to bottleneck states by using the extracted representations to determine grasp points and trajectories. These actions are representation-dependent and designed to efficiently guide the object toward bottleneck states.

\textbf{Exploration Action} ($a_e$) physically manipulates the object to transition unrecognizable states into recognizable ones. This exploration action is specifically designed for cases where the robot recognizes the object's presence but fails to comprehend its specific configuration.
Exploration Action is a complement to a selected recognition algorithm. Moreover, does not depend on an extracted incorrect abstract representation, but rather on the more basic properties (e.g. point cloud).

The key insight is that accurately recognizing the current state is not essential. Physical interactions can increase the likelihood of reaching recognizable configurations through generic manipulation and reduce self-occlusion.

\subsection{System Architecture}

Fig. \ref{fig:architecture} shows our system architecture, which comprises four main components: (1) Object Recognition Module, (2) Action Decision Policy, (3) Action Primitives, and (4) Behavior Cloning Model for task completion. 

\subsubsection{Object Recognition Module (ORM)}

This module extracts abstract representations from observations, which are used to execute Preparation actions. The partition between $\mathcal{R}$ and $\bar{\mathcal{R}}$ depends on the choice of an extraction method.

For rope manipulation, we use a graph-based representation algorithm \cite{choi2024mBEST}, which includes representation of the endpoints of the rope. For cloth manipulation, we implement a custom keypoint detector that identifies adjacent corners. 

This module is not required to recognize all possible object states, but must only succeed for states reachable via Exploration Action. This relaxed requirement renders the approach more practical for DOM, where achieving universal recognition is infeasible.

\begin{figure}[t]
\centering
    \begin{minipage}[b]{0.24\columnwidth}
        \includegraphics[width=0.98\columnwidth]{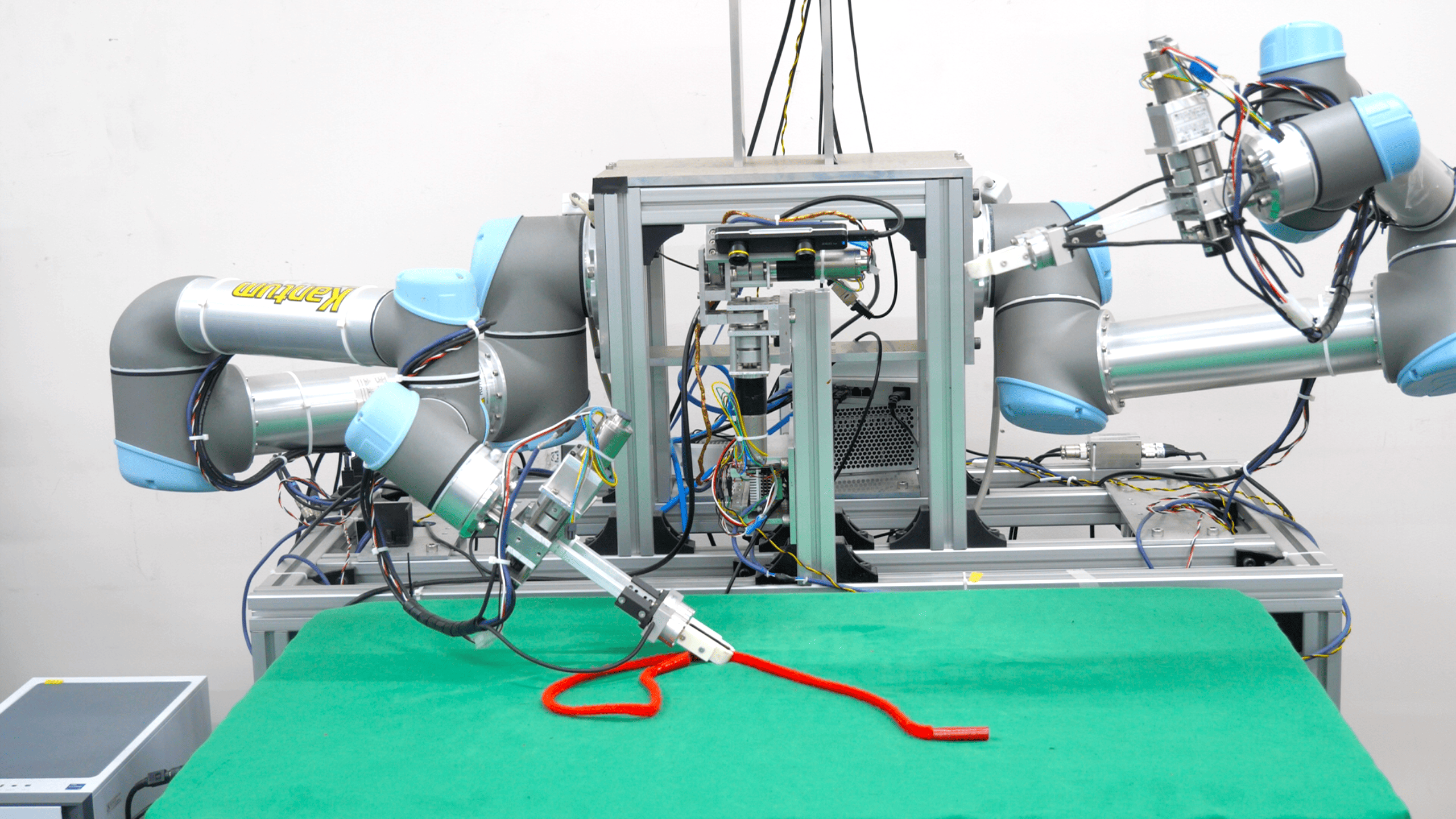}
    \end{minipage}
   \begin{minipage}[b]{0.24\columnwidth}
        \includegraphics[width=0.98\columnwidth]{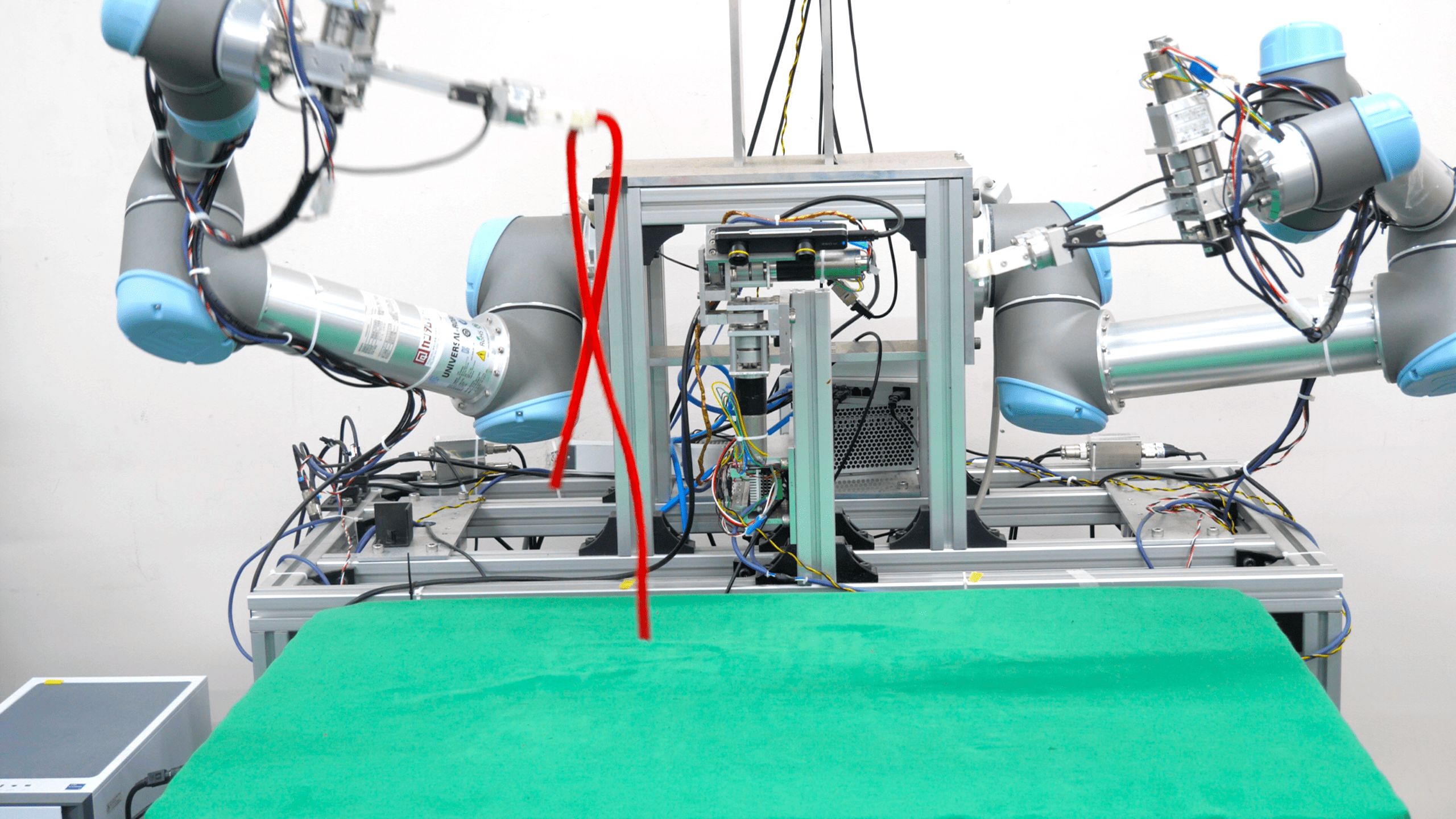}
    \end{minipage}
    \begin{minipage}[b]{0.24\columnwidth}
        \includegraphics[width=0.98\columnwidth]{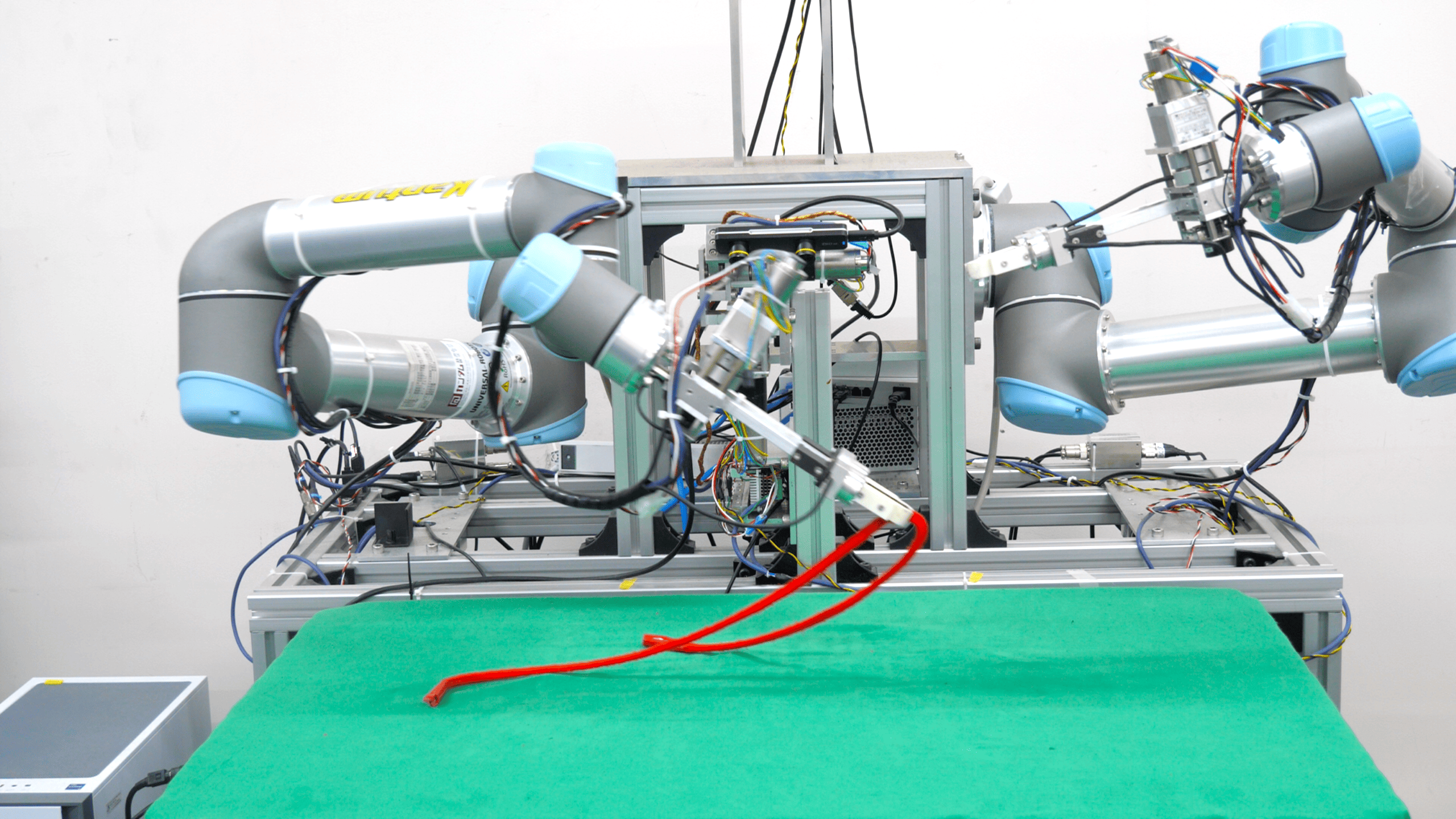}
    \end{minipage}
   \begin{minipage}[b]{0.24\columnwidth}
        \includegraphics[width=0.98\columnwidth]{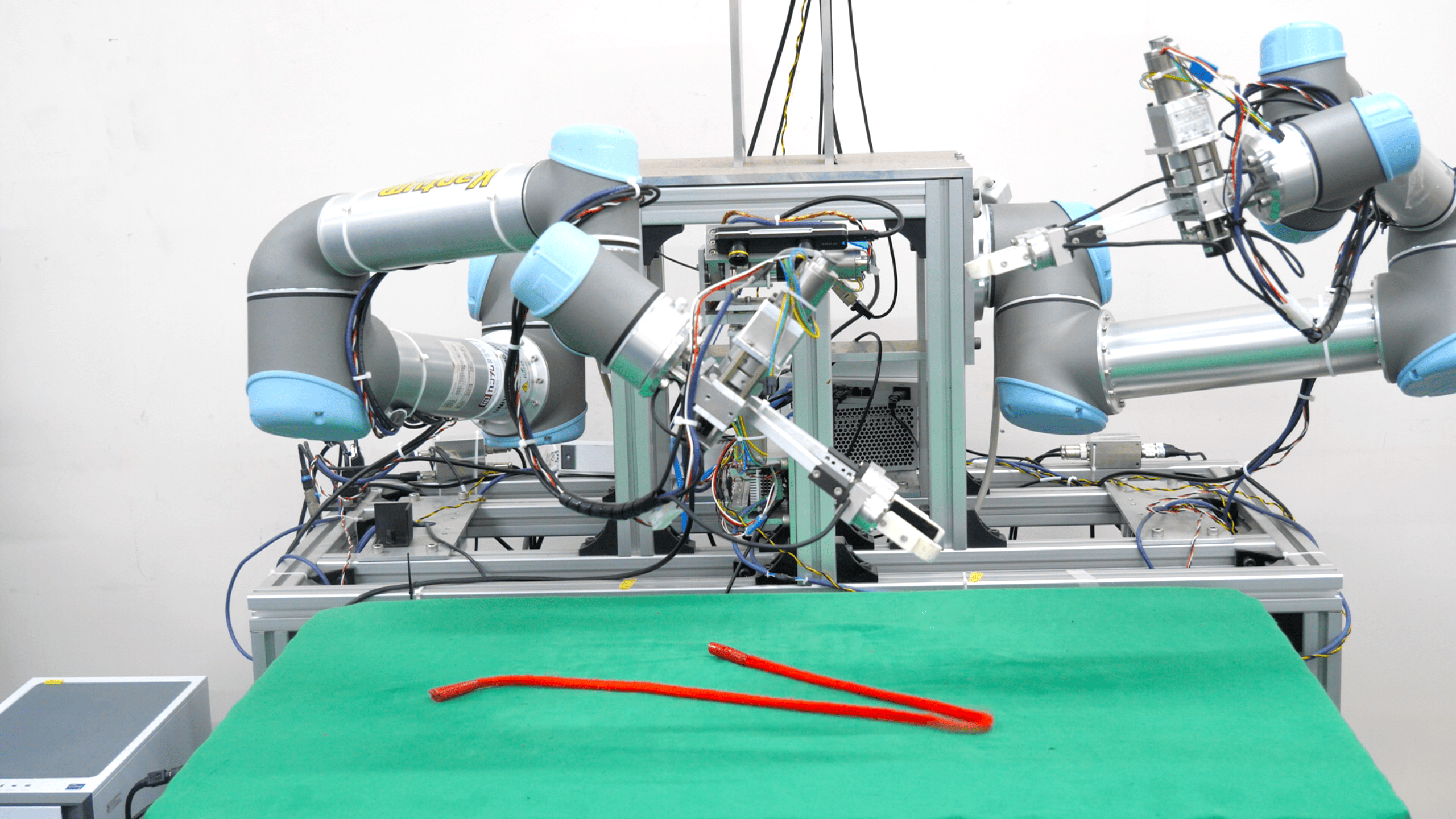}
    \end{minipage}

    \vspace{2mm}
    
    \begin{minipage}[b]{0.24\columnwidth}
        \includegraphics[width=0.98\columnwidth]{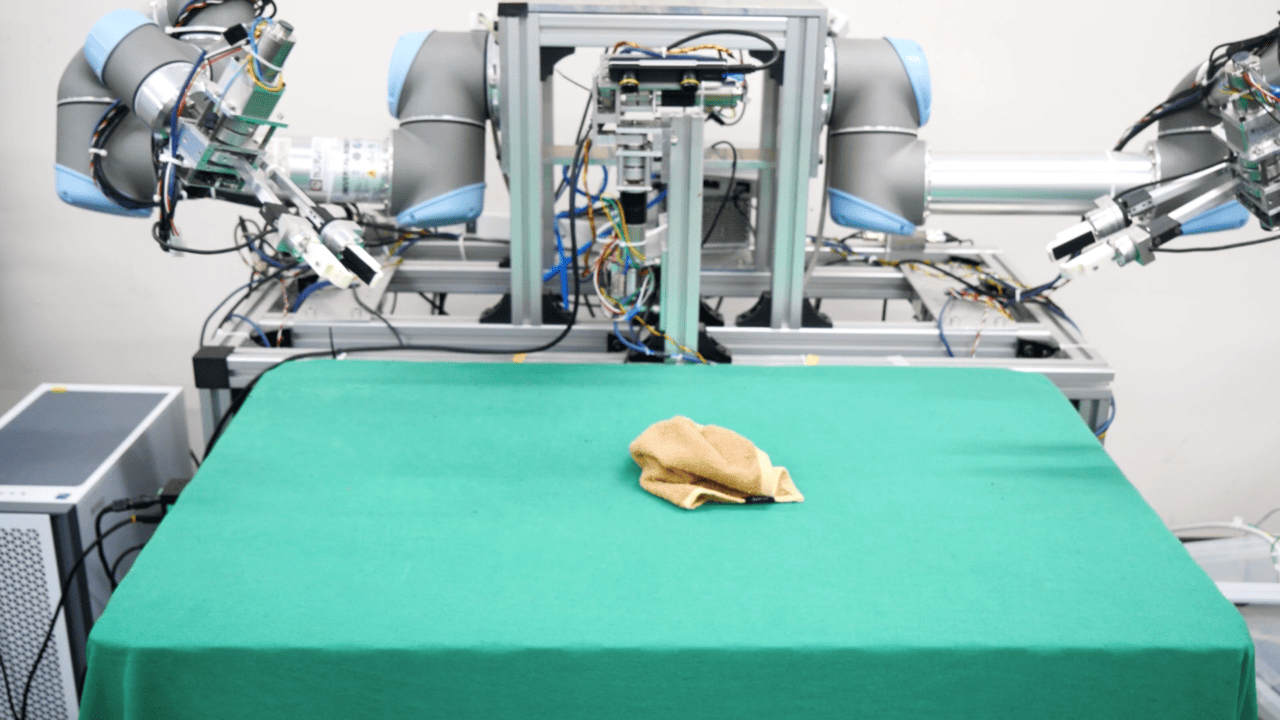}
    \end{minipage}
   \begin{minipage}[b]{0.24\columnwidth}
        \includegraphics[width=0.98\columnwidth]{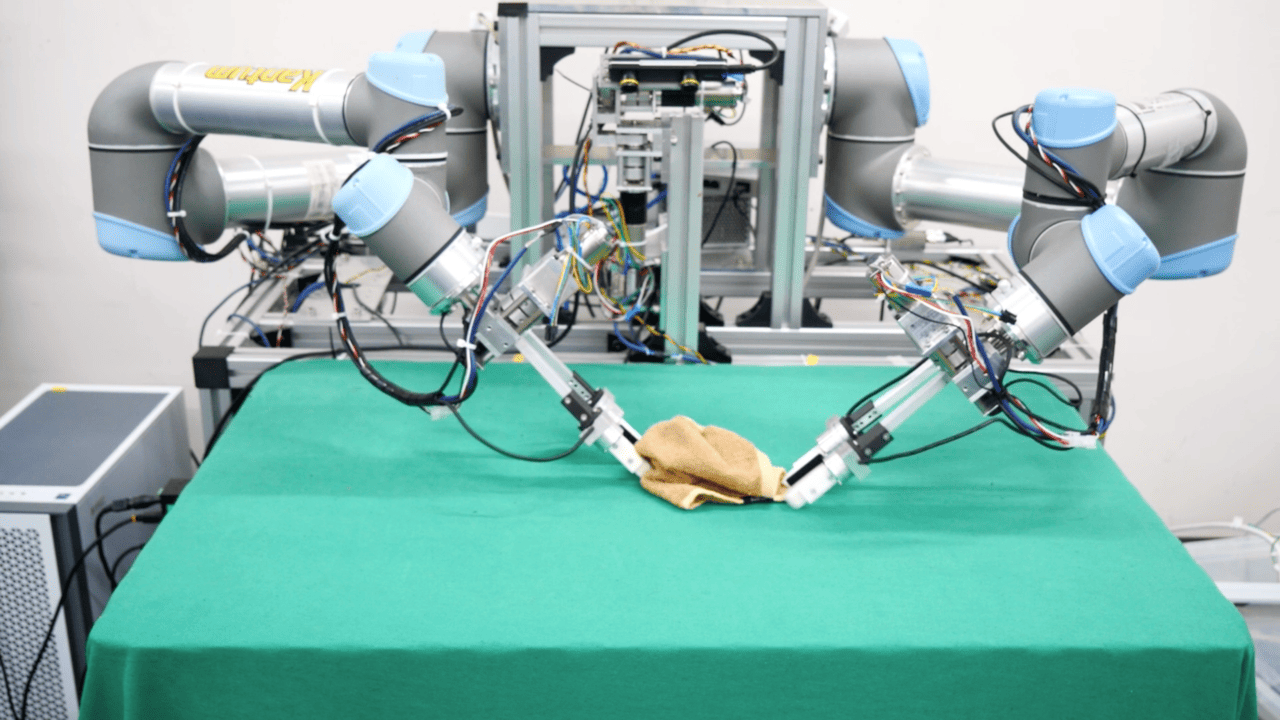}
    \end{minipage}
    \begin{minipage}[b]{0.24\columnwidth}
        \includegraphics[width=0.98\columnwidth]{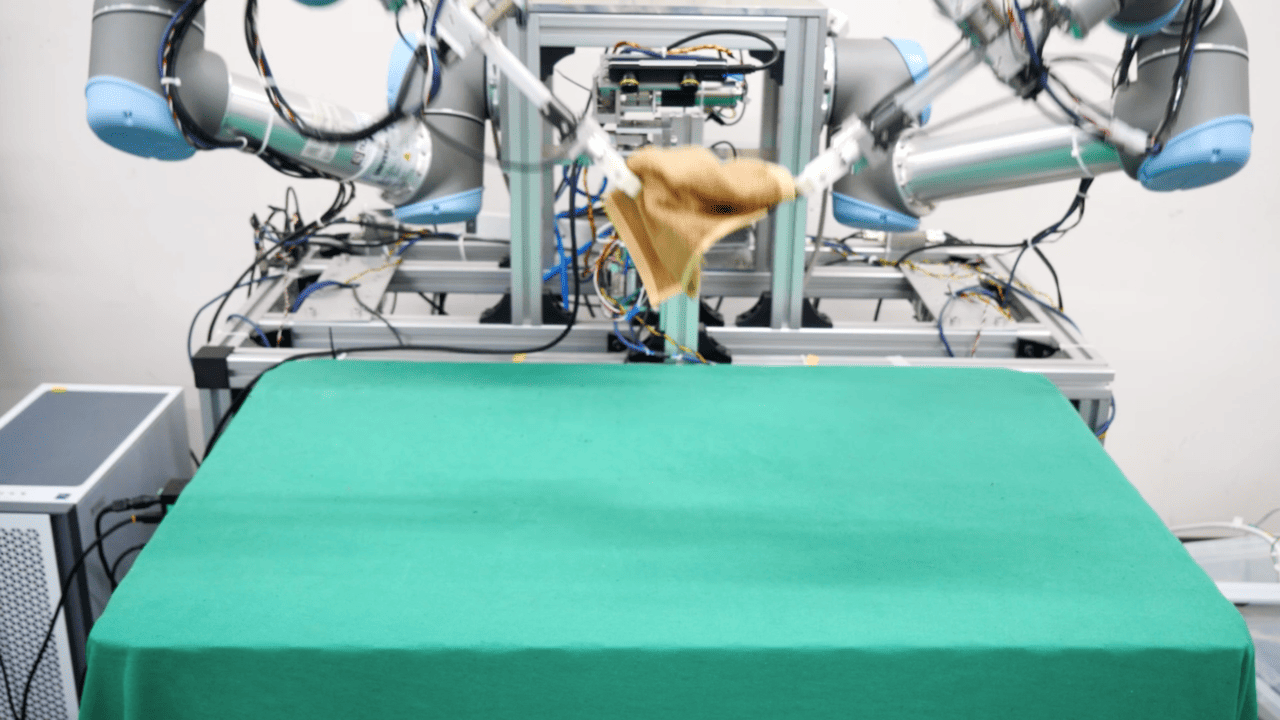}
    \end{minipage}
   \begin{minipage}[b]{0.24\columnwidth}
        \includegraphics[width=0.98\columnwidth]{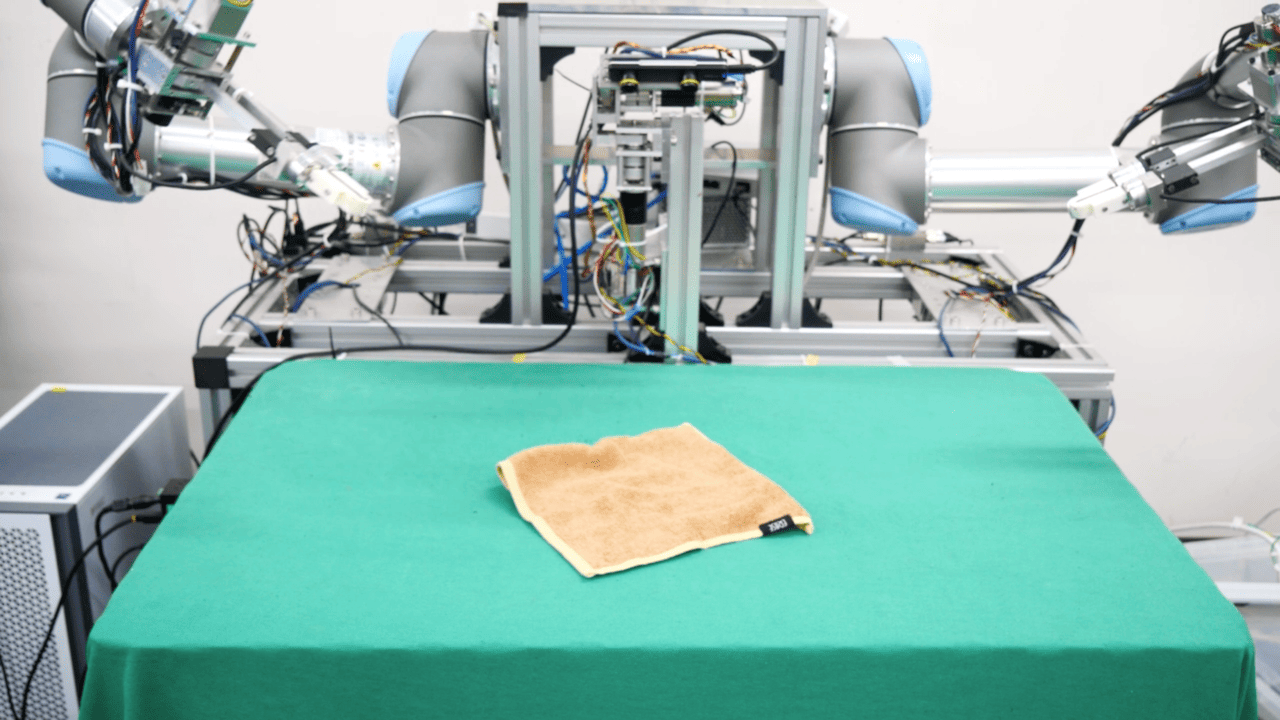}
    \end{minipage}
    \caption{Exploration Action of rope and cloth manipulation}
    \label{fig:exp_cloth}
\end{figure}

\begin{figure}[t]
\centering
    \begin{minipage}[b]{0.24\columnwidth}
        \includegraphics[width=0.98\columnwidth]{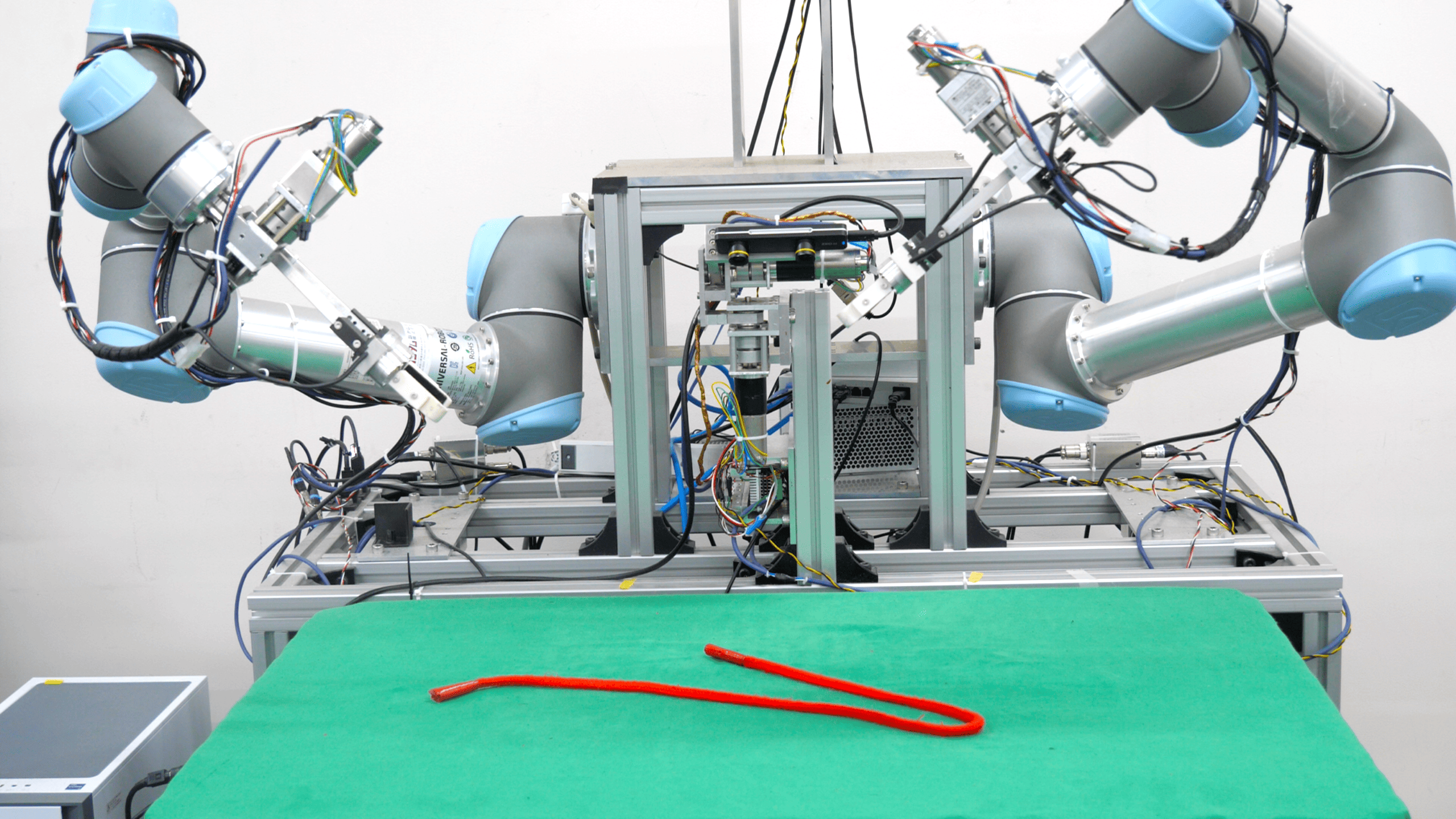}
    \end{minipage}
   \begin{minipage}[b]{0.24\columnwidth}
        \includegraphics[width=0.98\columnwidth]{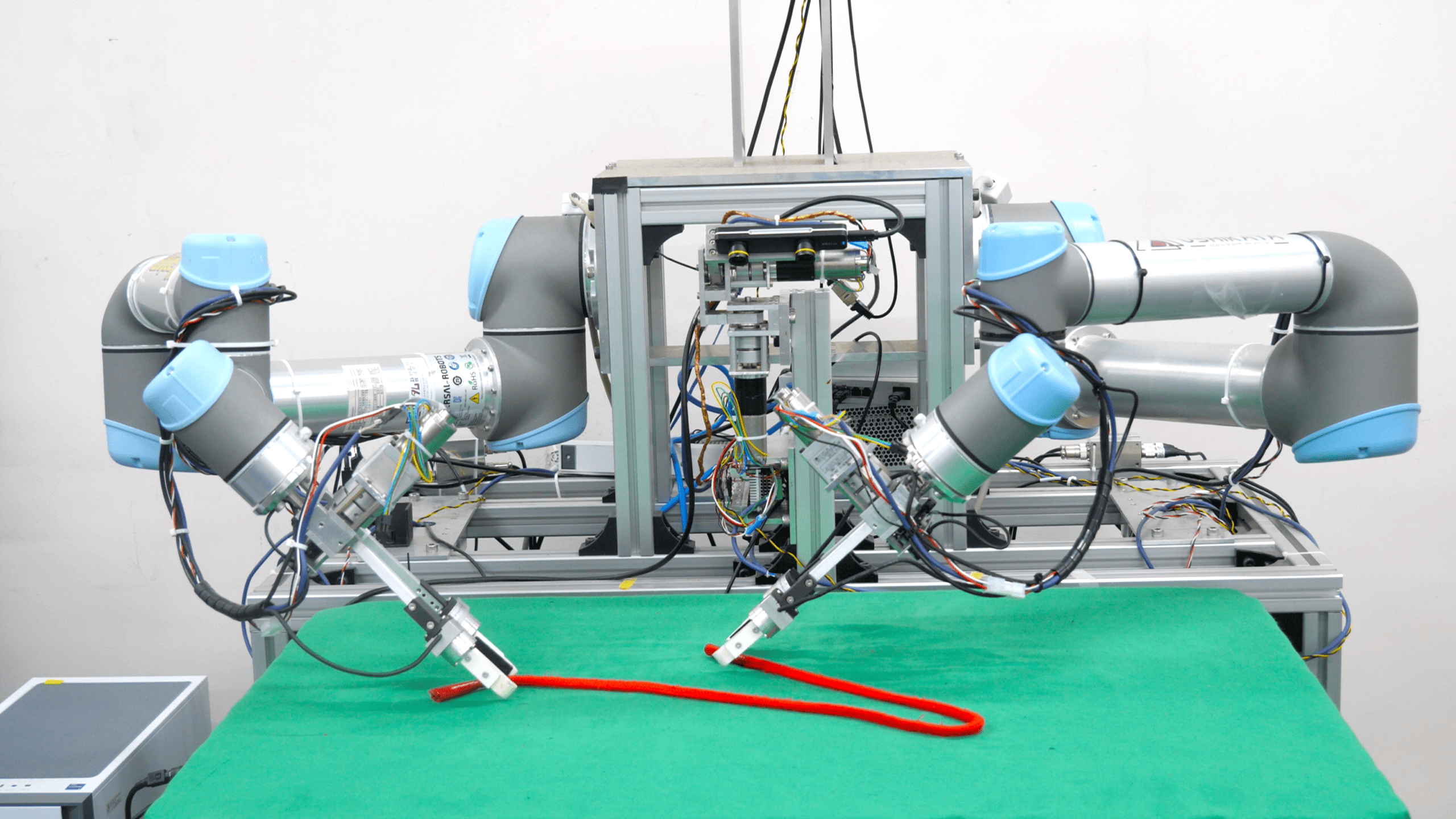}
    \end{minipage}
    \begin{minipage}[b]{0.24\columnwidth}
        \includegraphics[width=0.98\columnwidth]{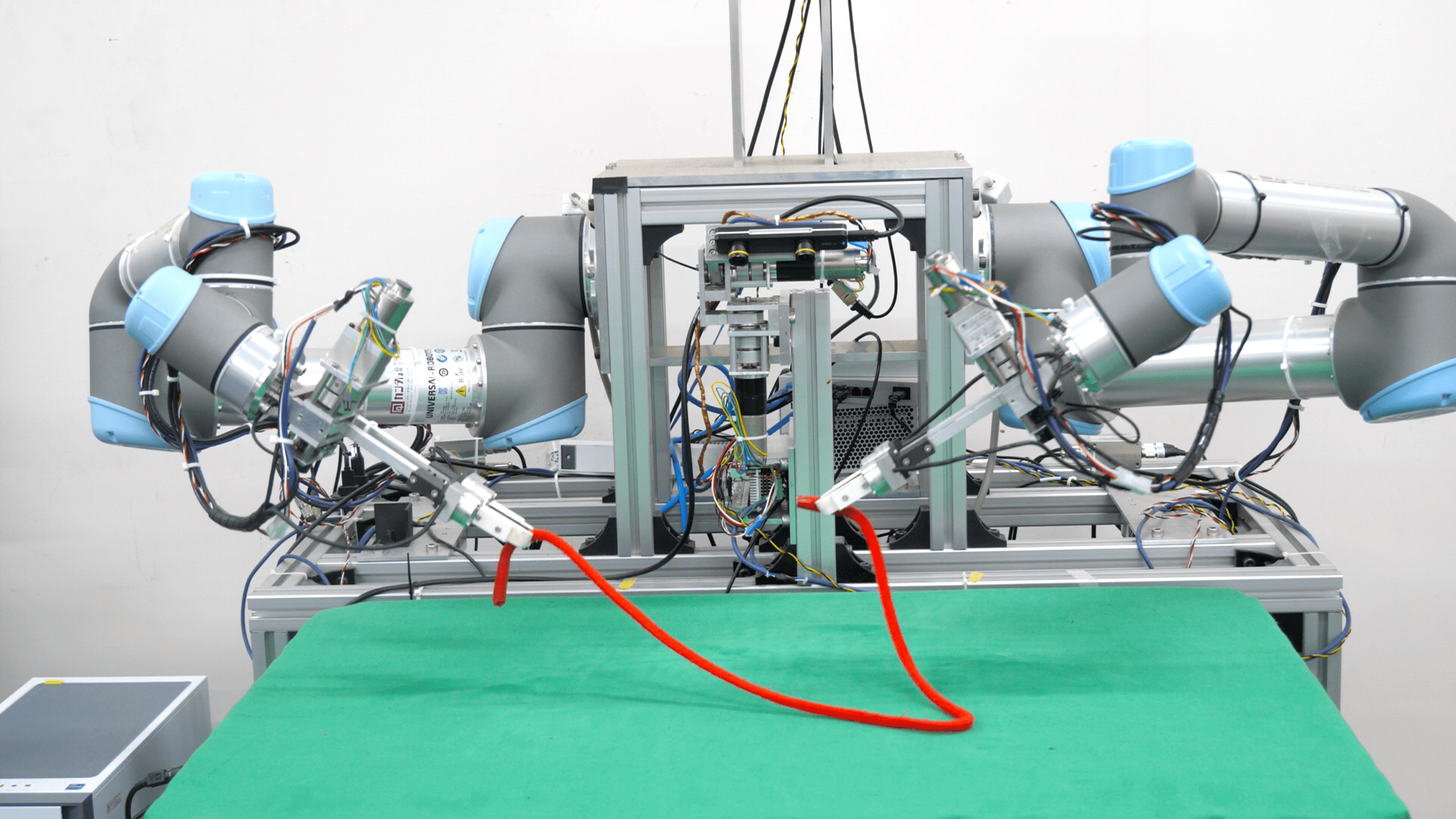}
    \end{minipage}
   \begin{minipage}[b]{0.24\columnwidth}
        \includegraphics[width=0.98\columnwidth]{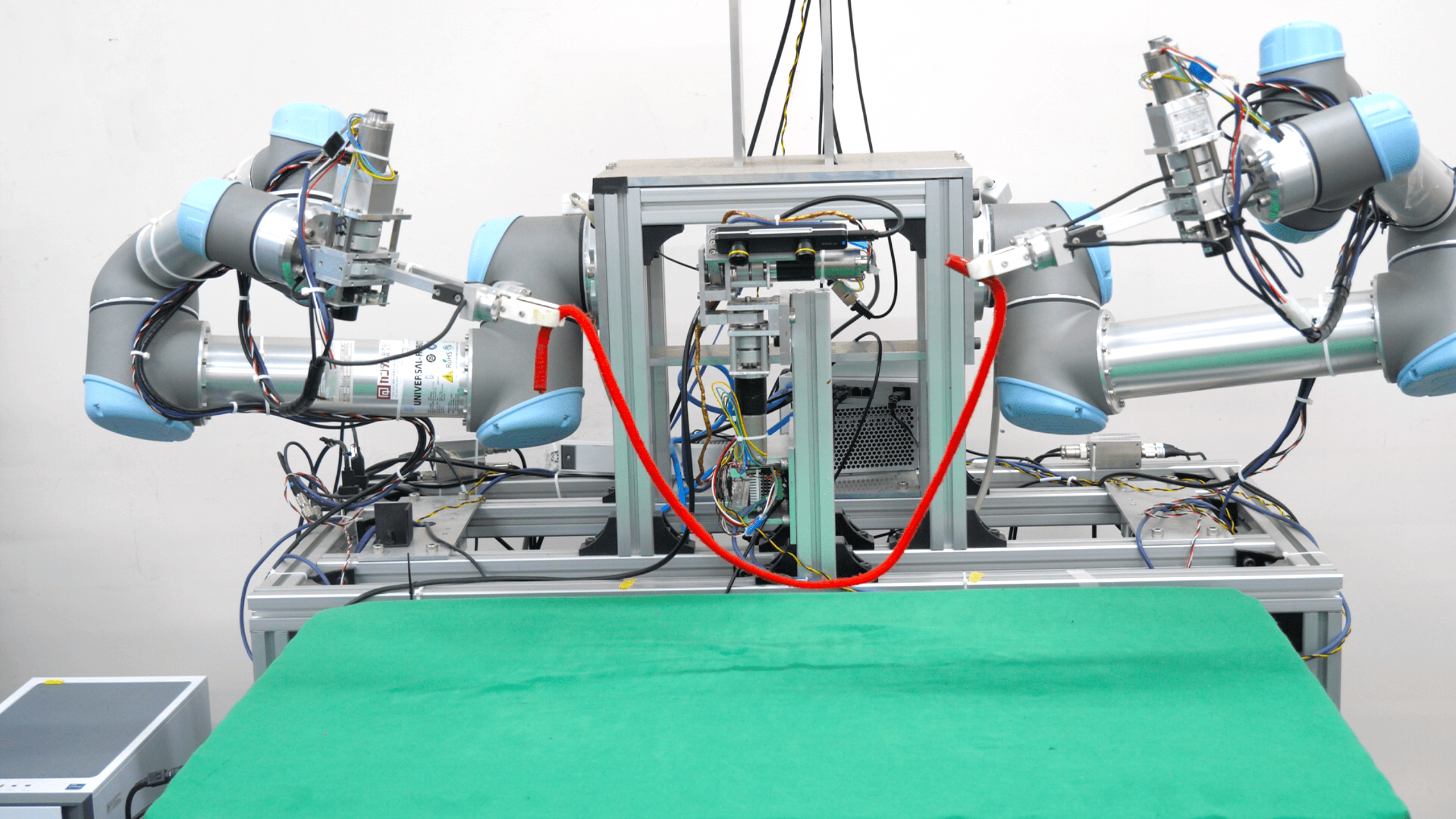}
    \end{minipage}

    \vspace{2mm}
    
    \begin{minipage}[b]{0.24\columnwidth}
        \includegraphics[width=0.98\columnwidth]{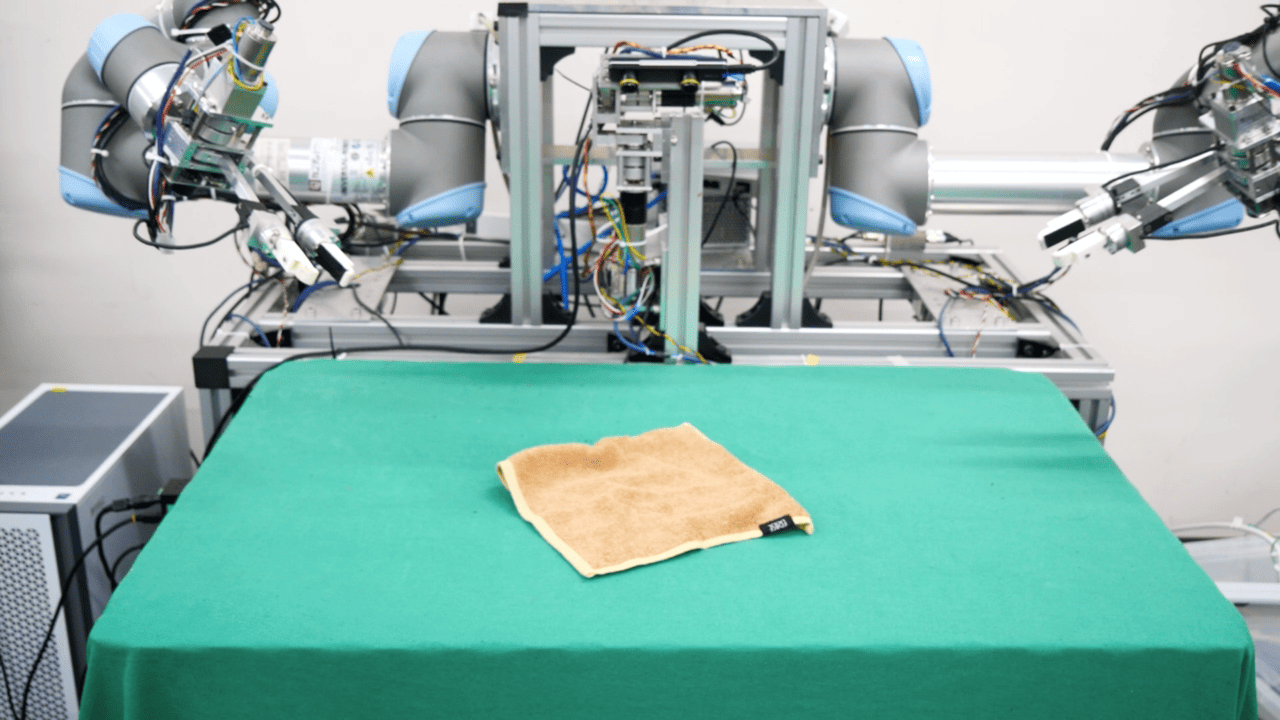}
    \end{minipage}
   \begin{minipage}[b]{0.24\columnwidth}
        \includegraphics[width=0.98\columnwidth]{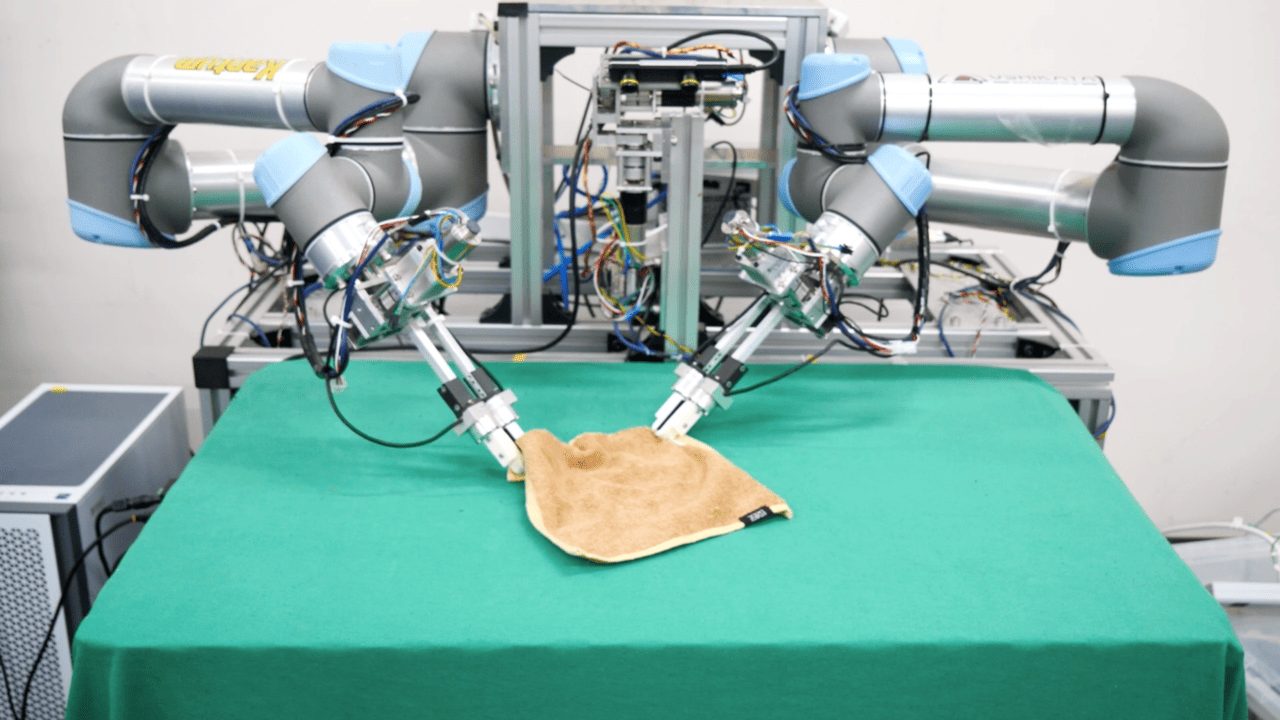}
    \end{minipage}
    \begin{minipage}[b]{0.24\columnwidth}
        \includegraphics[width=0.98\columnwidth]{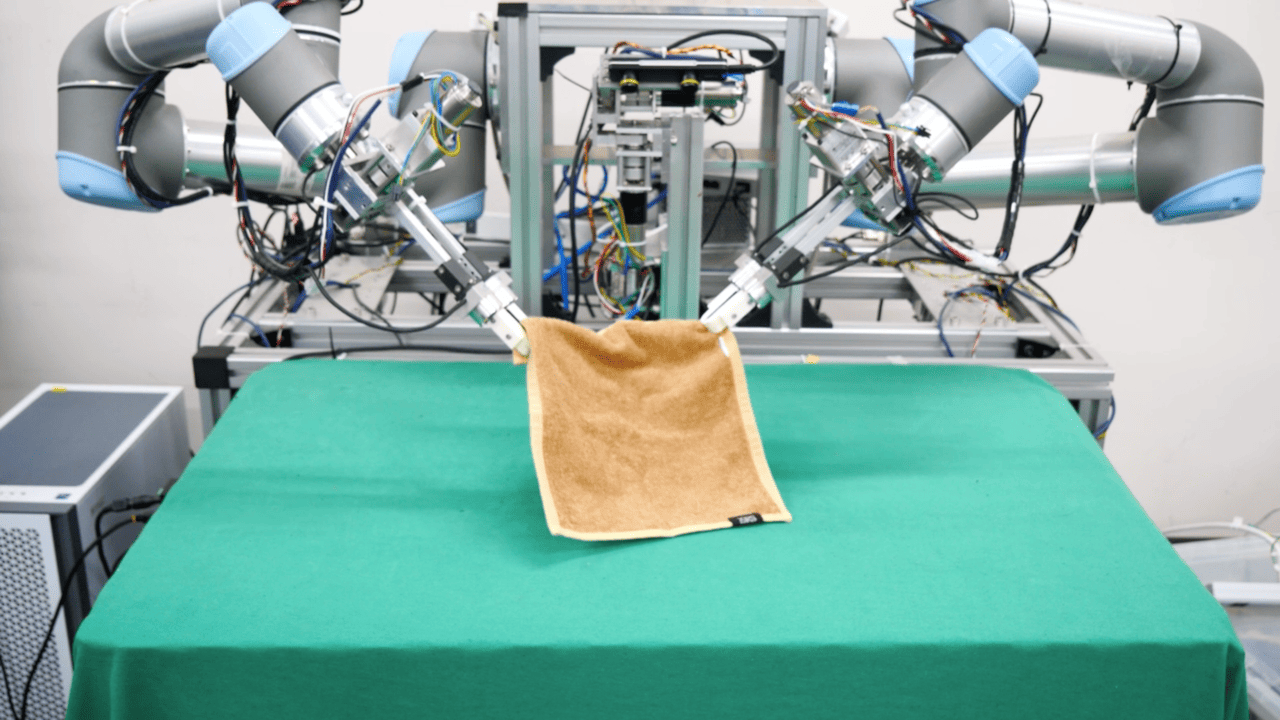}
    \end{minipage}
   \begin{minipage}[b]{0.24\columnwidth}
        \includegraphics[width=0.98\columnwidth]{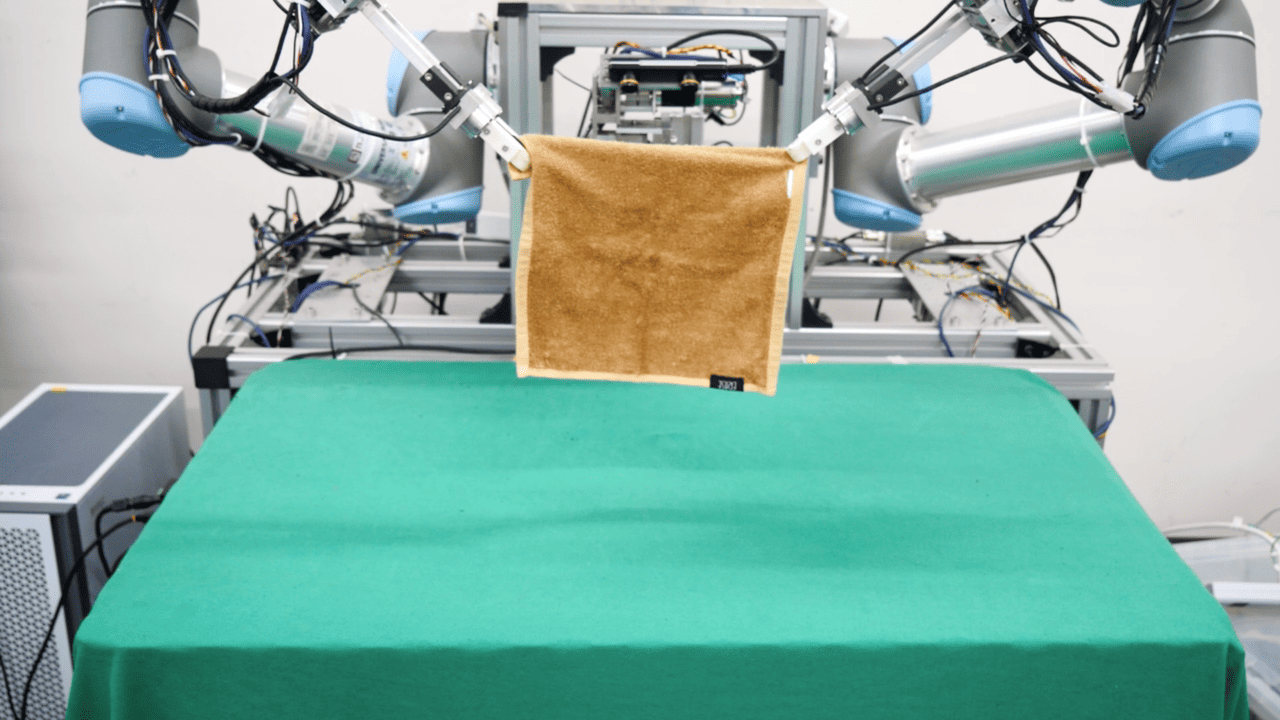}
    \end{minipage}
    \caption{Preparation Action of rope and cloth manipulation}
    \label{fig:pre_cloth}
    \vspace{-0.7em}
\end{figure}

\subsubsection{Action Decision Policy (ADP)}
This module classifies the current object configuration into either the recognizable states $\mathcal{R}$ or the unrecognizable states $\bar{\mathcal{R}}$, and selects the corresponding action primitive. To achieve this without explicit state labels, the system must autonomously distinguish between these states to enable the selection of a task-oriented primitive $a_p$ when recognition succeeds, or an exploration action $a_e$ when it fails.

In this study, we employed a GPT-5.2 \cite{openai2025gpt}, a vision-language model (VLM). The model is designed to perform reasoning tasks, such as those required by the Action Decision Policy, to enable state classification and action selection. The VLM evaluates recognition correctness by comparing RGB images with overlaid extracted representations (examples are shown in Fig.~\ref{fig:recog}). 
Fig.~\ref{fig:vlm_example} shows the general structure of prompts. The structure consists of four elements: (1) the verification objective defining the classification task, (2) explanations for input images, (3) domain-specific conditions that characterize correct representations (e.g., endpoint visibility for ropes, corner adjacency for cloth), and (4) a structured output format for explicit reasoning followed by a binary decision.
With these components, the ADP then assesses whether the extracted representation accurately reflects the object's physical structure and selects the appropriate action primitive $a \in \{a_p, a_e\}$ for the given state. 

\subsubsection{Action Primitives}

We implement two types of actions executed based on the Action Decision Policy:

\textbf{Exploration Action:} When the Action Decision Policy determines that recognition has failed ($s \in \bar{\mathcal{R}}$), the system executes $a_e$ without relying on abstract representations:

\begin{itemize}
    \item \textit{Rope}: Extract the point from the segmented object's point cloud that is closest to the camera. Grasp this point and execute a predefined lifting trajectory to alter the object's configuration and promote natural unfolding. The lifting motion is introduced to reduce self-occlusion by separating overlapping segments.
    
    \item \textit{Cloth}: Extract points at the maximum and minimum x-coordinates in the segmented point cloud. Bimanually grasp these points, lift the object, and release it into the air. This dynamic motion is introduced to utilize air resistance. Consequently, this leads to the unfolding of the cloth, exposing previously occluded regions, and guiding the object into recognizable states.
\end{itemize}

A selected Object Recognition Module has certain assumptions or limitations for successful recognition (e.g., most segments are visible, an object approximates a semi-planar configuration). The design of Exploration Action $a_e$ is aimed at restoring the operational assumptions of the Object Recognition Module by physically altering the object's configuration. Specifically, $a_e$ is not just a random movement, but as a disentanglement process designed for simplifying the object's geometry until it satisfies the specific input constraints required for successful recognition in $\mathcal{R}$.

\begin{figure}[t]
\centering
    \begin{minipage}[b]{0.38\columnwidth}
        \includegraphics[width=0.98\columnwidth]{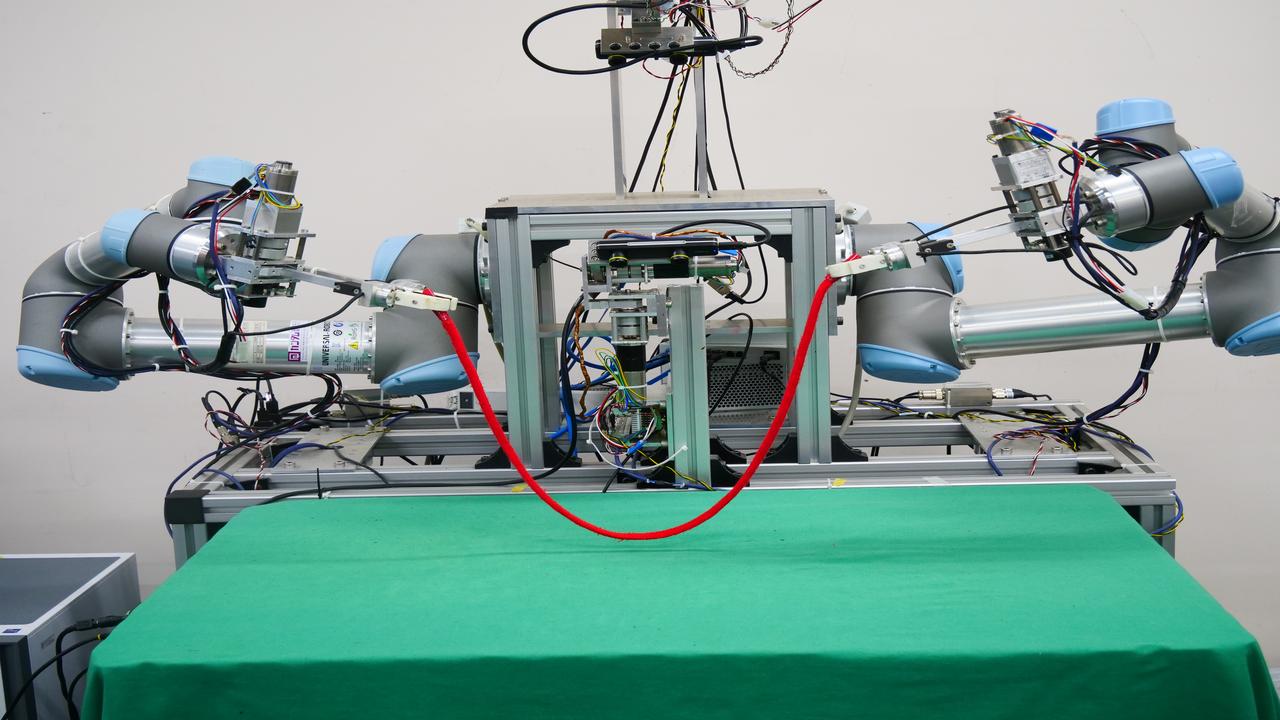}
    \end{minipage}
    \begin{minipage}[b]{0.38\columnwidth}
        \includegraphics[width=0.98\columnwidth]{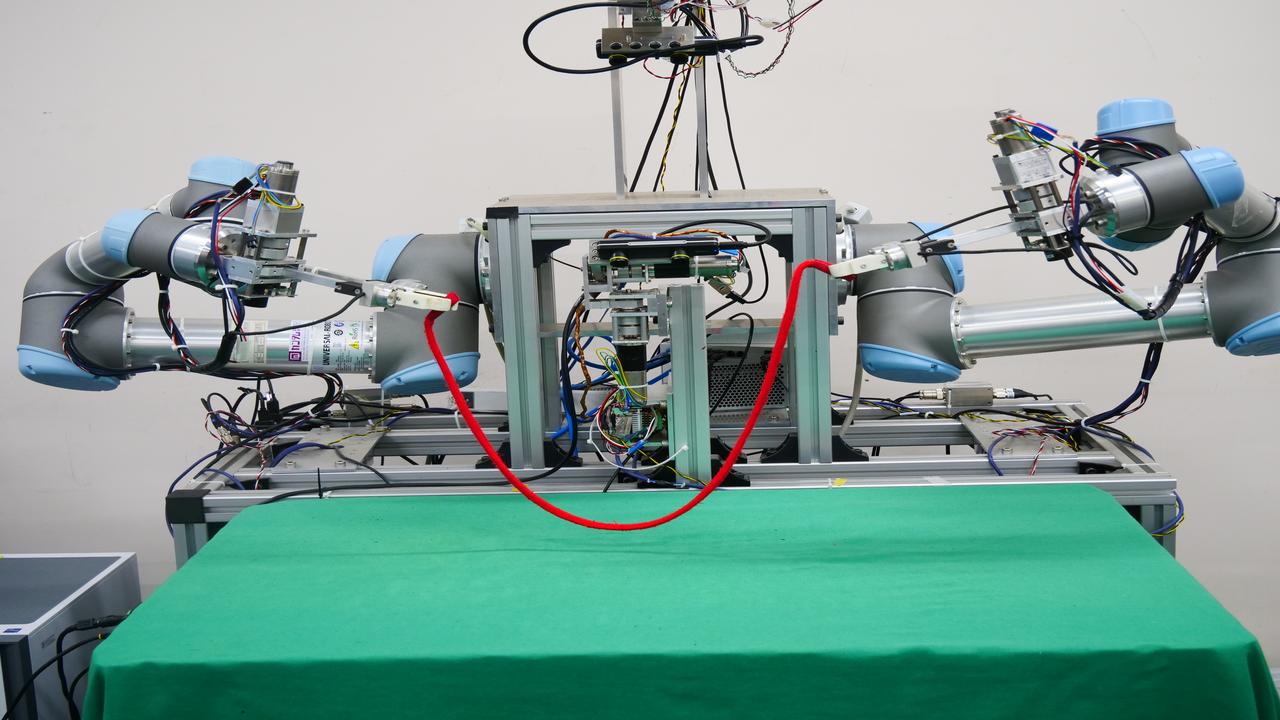}
    \end{minipage}

    \vspace{2mm}
    
    \begin{minipage}[b]{0.38\columnwidth}
        \includegraphics[width=0.98\columnwidth]{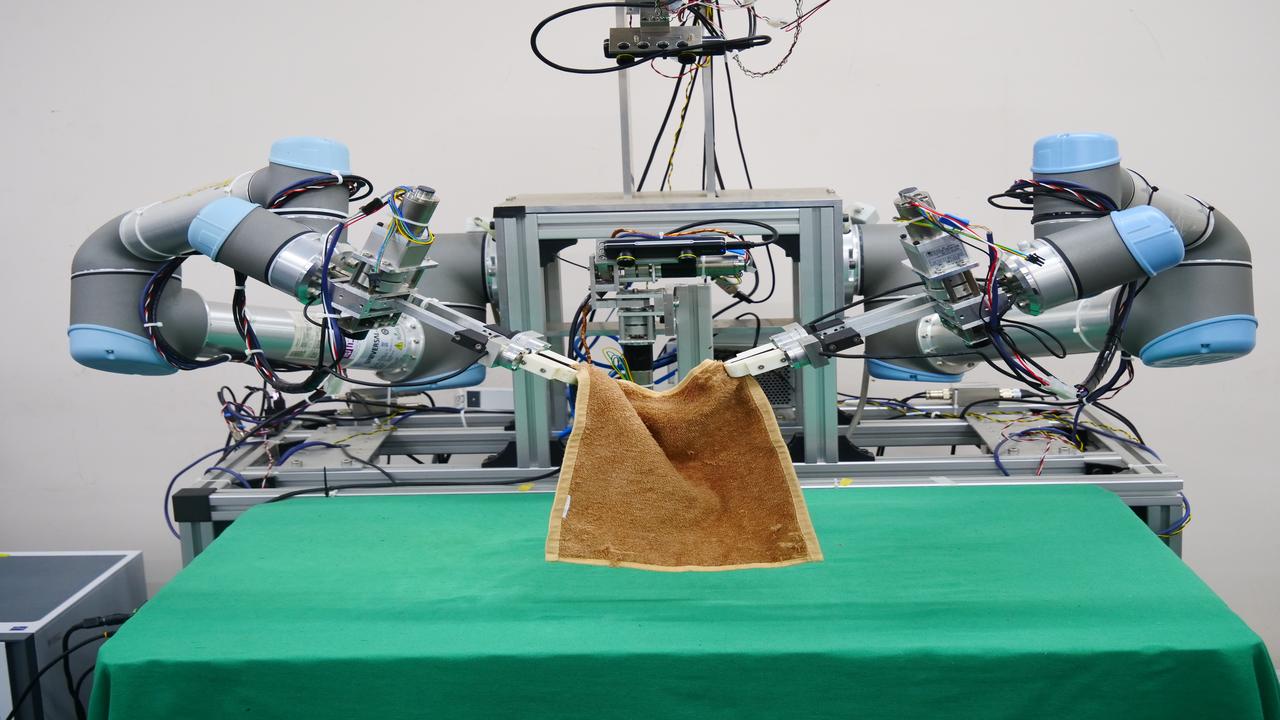}
    \end{minipage}
    \begin{minipage}[b]{0.38\columnwidth}
        \includegraphics[width=0.98\columnwidth]{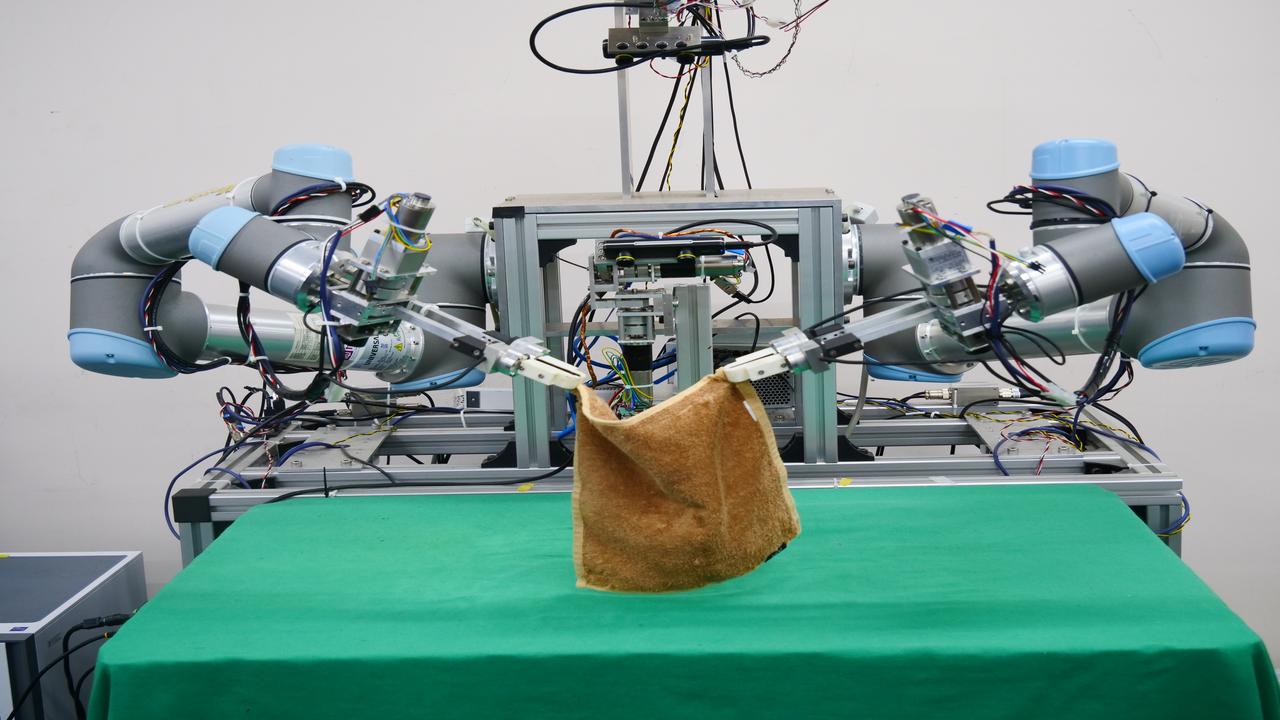}
    \end{minipage}
    \caption{Examples of bottleneck states for rope and cloth manipulation}
    \label{fig:example_bottleneck}
    \vspace{-0.7em}
\end{figure}

\textbf{Preparation Action:} When the Action Decision Policy judges that recognition has succeeded ($s \in \mathcal{R}$), the system executes $a_p$ using the extracted representation:

\begin{itemize}
    \item \textit{Rope}: Extract 3D point cloud coordinates of graph endpoints. Compute the gripper pose aligned with the edge direction near the endpoints. Grasp both endpoints and alter the object’s configuration to a bottleneck state (examples are shown in Fig.~\ref{fig:example_bottleneck}).
    
    \item \textit{Cloth}: Extract 3D coordinates of detected adjacent corners. Compute gripper poses aligned with directions from corners toward the object center. Bimanually grasp the corners and alter the object’s configuration to a bottleneck state (examples are shown in Fig.~\ref{fig:example_bottleneck}). 
\end{itemize}
A Preparation Action is considered successful if both grippers grasp the extracted keypoints, and all successful resulting object configurations are considered to be within $\mathcal{S}_b$.

\subsubsection{Behavior Cloning (BC) Model}
After transitioning the object to a bottleneck state $s_b$, the system conducts its tasks with the learned policy. In this study, we adopt the Action Chunking with Transformer (ACT) \cite{zhao2023learning} for implementation.
As described in Sec.~\ref{subsubsec:bc_model}, learning from bottleneck states enables data-efficient traing while mitigating the variability in object deformation.

\subsection{Overall Architecture}

Algorithm~\ref{alg:system} summarizes the complete system flow. Given an arbitrary OOD initial state $s_\mathrm{out}$, the system first processes the input image through the Object Recognition Module (ORM), which produces an overlay image visualizing the extracted keypoint representation. Subsequently, both the raw input image and the overlay image are passed to the Action Decision Policy (ADP) to determine whether the current state is in recognizable states. If the system judges $s \in \bar{\mathcal{R}}$, an Exploration Action $a_e$ is executed to physically reshape the object and try to restore recognition. If the system judges $s \in \mathcal{R}$, a Preparation Action $a_p$ is executed to transition the object into a bottleneck state $s_b$, from which the BC model completes the tasks.
In this study, only the BC Model is trained using demonstrations from bottleneck states, and the Exploration Action and Preparation Action are not learned in this study.

\section{EXPERIMENT}
\label{sec:experiment}

To evaluate the effectiveness of our proposed system for handling OOD states in DOM, we conducted experiments on two distinct manipulation tasks: knot tying and cloth folding. This section discusses the experimental setup and results.

\begin{algorithm}[t]
    \caption{Overall system flow of ExBot}
    \label{alg:system}
    
    \SetKwInOut{Input}{Input}
    \SetKwInOut{Notation}{Notation}
    \Input{OOD State $s_{out}$}
    
    \BlankLine
    
    $s \leftarrow s_{out}$ \;
    $o \leftarrow GetImage(s)$ \;
    $o_{rep} \leftarrow ORM(o)$ \;
    
    \BlankLine
    \While{$ADP(o, o_{rep})$ judges $s \in \bar{\mathcal{R}}$}{
        $s \leftarrow a_e(s)$ \;
        $o \leftarrow GetImage(s)$ \;
        $o_{rep} \leftarrow ORM(o)$ \;
    }
    
    \BlankLine
    $s_{b} \leftarrow a_p(s)$ \;
    
    \BlankLine
    Execute policy $\pi_{\text{BC}}(s_{b})$ \;
    
\end{algorithm}

\subsection{Task Definition and Data Collection}
\label{subsec:task_def_data_collect}
We defined two manipulation tasks for the Behavior Cloning (BC) model to learn:

\textbf{Knot Tying:}
The objective is to tie a knot in a red rope starting from various unknotted OOD states. For this task, we collected 157 demonstration episodes starting from bottleneck states, defined as the state where both ends of the rope are grasped by the robot's grippers (see Fig. \ref{fig:example_bottleneck}). The task is considered successful if an overhand knot is created.

\textbf{Cloth Folding:} 
The objective is to fold a beige cloth starting from rumpled states. We collected 138 demonstration episodes starting from bottleneck states, where the robot grasps two adjacent corners of the cloth (Fig. \ref{fig:example_bottleneck}). The task is deemed successful if the system executes two successive folding motions to reach a four-layered configuration, regardless of the precise alignment of the edges.

In this study, an episode is considered a failure when the number of Exploration Actions reached 20. Analysis revealed that such episodes converge to stable states (characterized by minimal state changes and consistent ADP judgments) before reaching this limit, supporting 20 as an appropriate threshold.

\subsection{OOD State Creation Protocol}
\begin{figure}[t]
\centering
    \begin{minipage}[b]{0.28\columnwidth}
        \includegraphics[width=0.98\columnwidth]{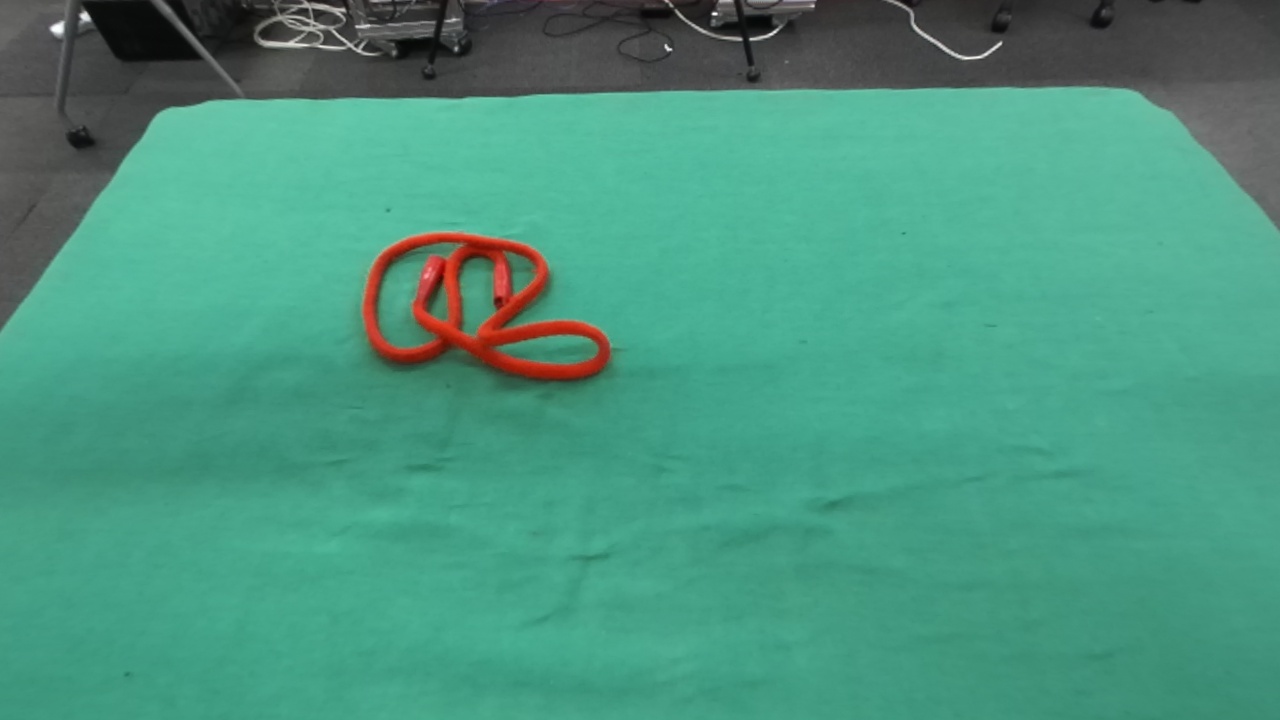}
    \end{minipage}
    \begin{minipage}[b]{0.28\columnwidth}
        \includegraphics[width=0.98\columnwidth]{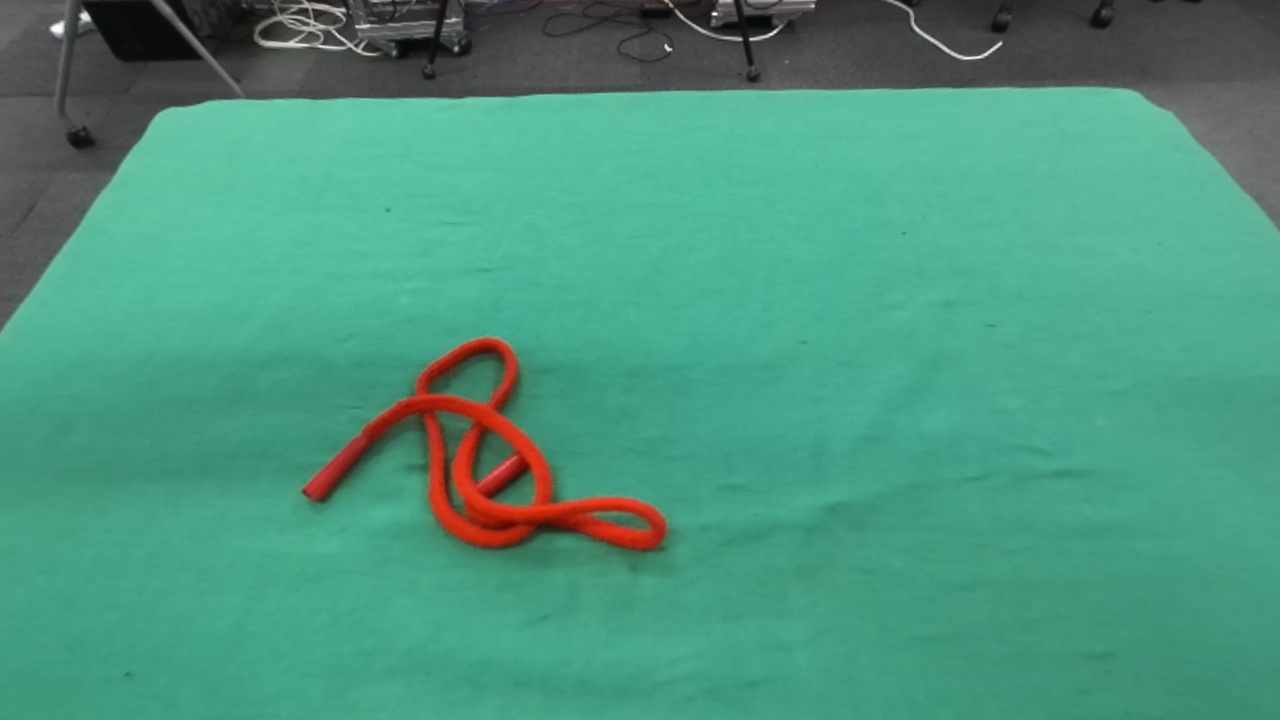}
    \end{minipage}
    \begin{minipage}[b]{0.28\columnwidth}
        \includegraphics[width=0.98\columnwidth]{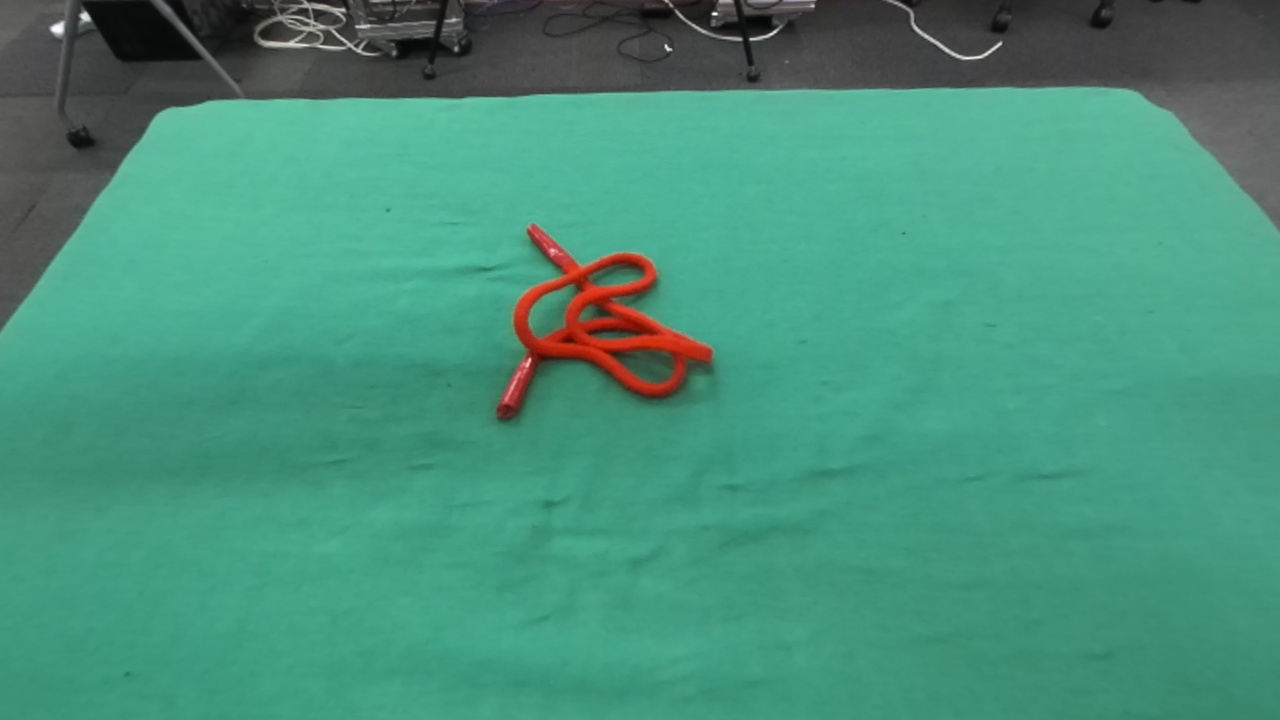}
    \end{minipage}

    \vspace{2mm}
    
    \begin{minipage}[b]{0.28\columnwidth}
        \includegraphics[width=0.98\columnwidth]{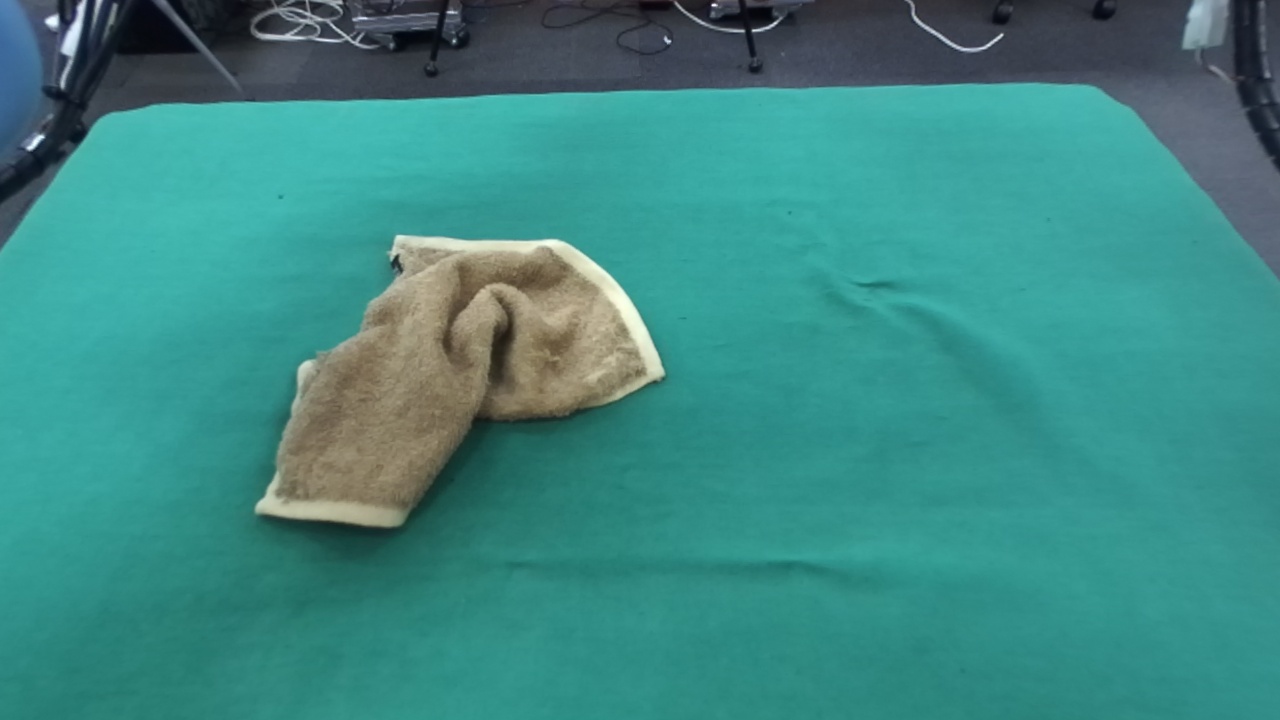}
    \end{minipage}
    \begin{minipage}[b]{0.28\columnwidth}
        \includegraphics[width=0.98\columnwidth]{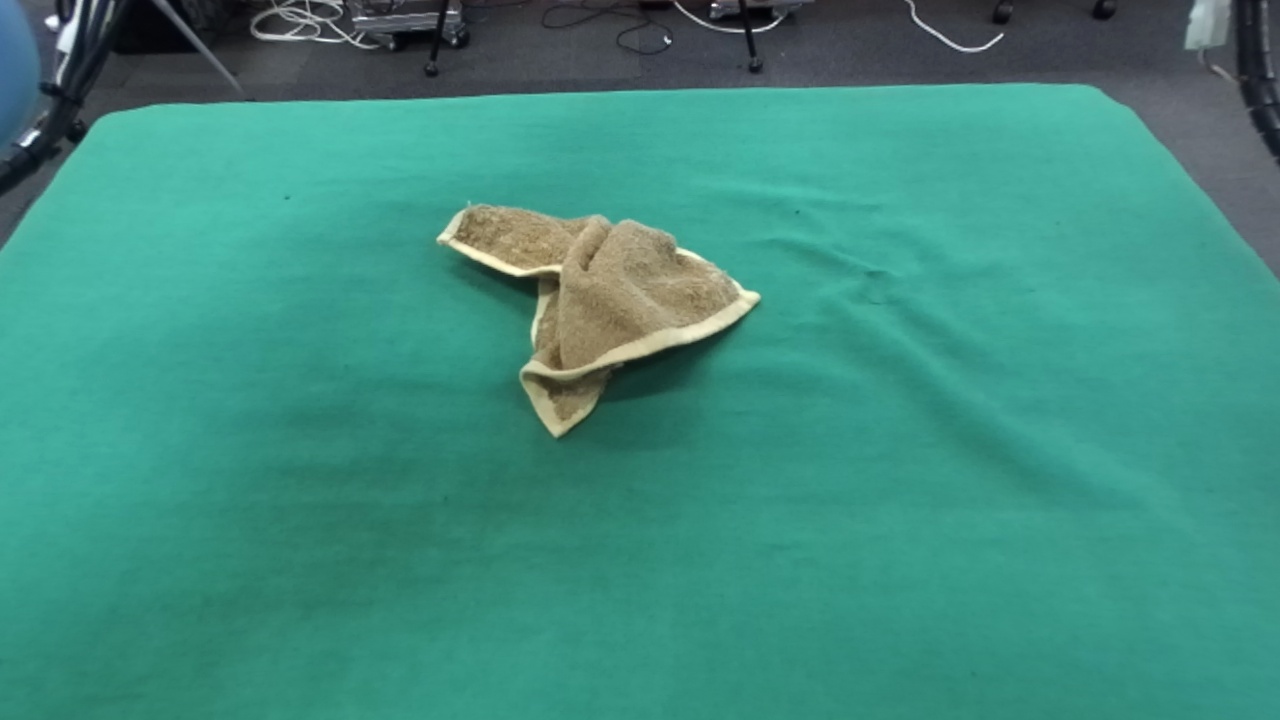}
    \end{minipage}
    \begin{minipage}[b]{0.28\columnwidth}
        \includegraphics[width=0.98\columnwidth]{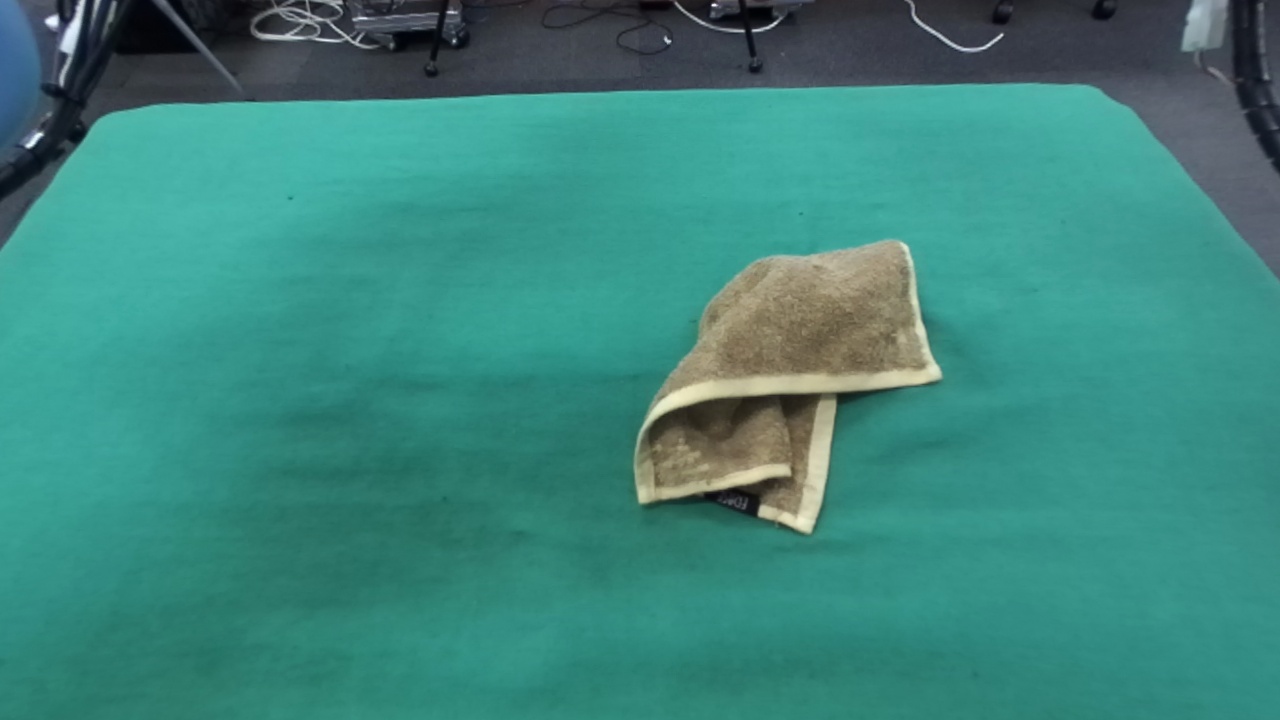}
    \end{minipage}
    \caption{OOD state examples created based on OOD protocol. Initial OOD states created by this protocol have positional diversity and complexity.}
    \label{fig:example_ood}
\end{figure}

To systematically evaluate the system's robustness to OOD initial conditions, we established protocols for creating diverse OOD states for both tasks. Fig. \ref{fig:example_ood} illustrates examples of OOD states.

\textbf{Rope Manipulation OOD Protocol:}
We created distinct OOD initial states for the rope following a standardized procedure:
\begin{enumerate}
    \item Unknot pre-existing knots in the rope.
    \item Fold the rope in half to shorten its effective length.
    \item Hold the folded rope in one hand.
    \item Throw it onto the table surface.
\end{enumerate}
In addition, none of the initial configurations contained pre-existing knots.

\textbf{Cloth Manipulation OOD Protocol:}
We generated distinct OOD initial states for the cloth following a similar standardized procedure:
\begin{enumerate}
    \item Roll up the cloth into a crumpled state.
    \item Hold the crumpled cloth in one hand.
    \item Throw it onto the table surface.
\end{enumerate}

We emphasize that the throwing motion intentionally introduced severe self-occlusion and high configurational variability in the initial states. Moreover, it replicated harsh real-world conditions that are challenging for standard perception systems.

\begin{figure}[t]
\centering
    \begin{minipage}[b]{0.43\columnwidth}
        \includegraphics[width=0.98\columnwidth]{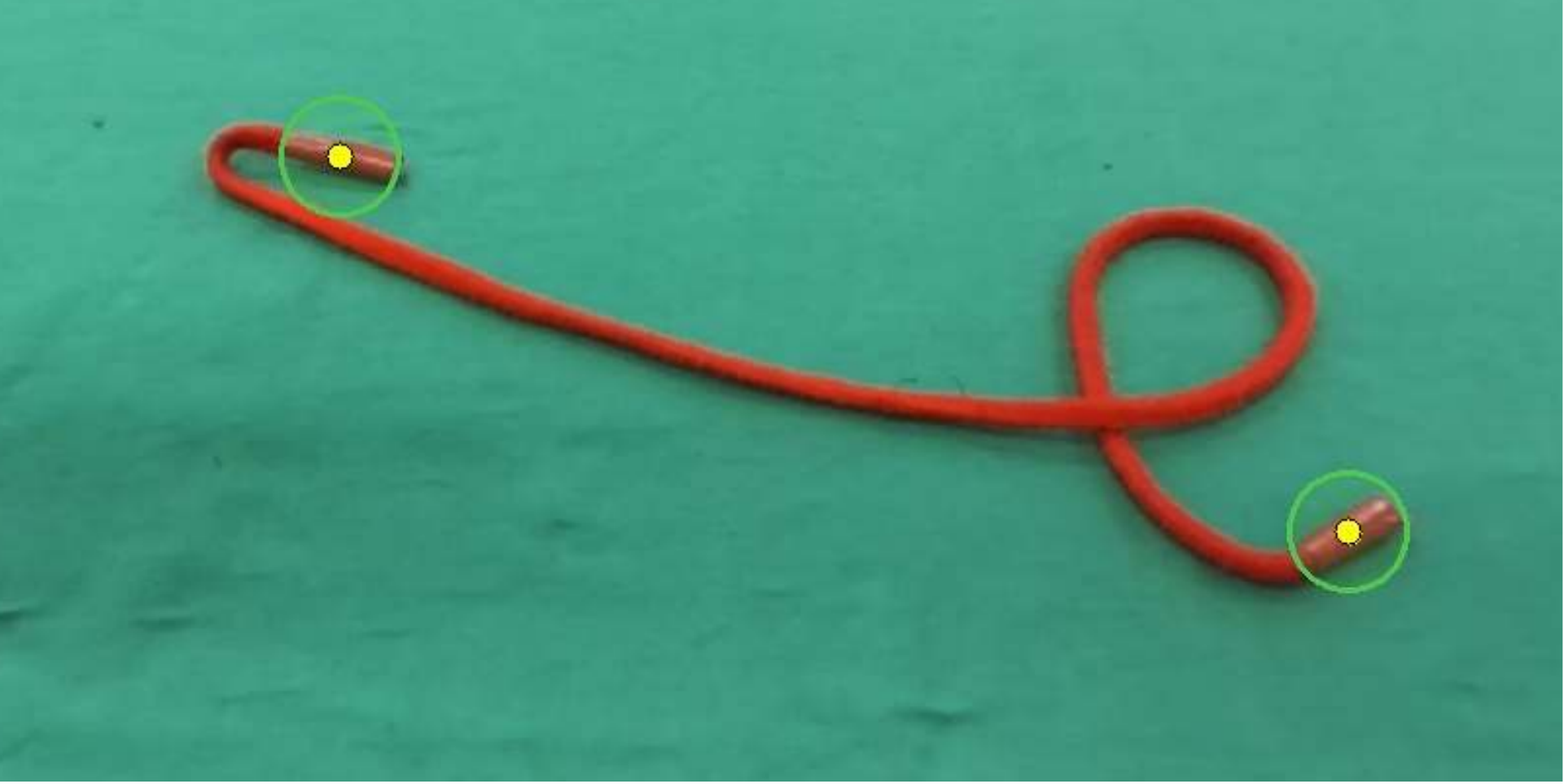}
    \end{minipage}
    \begin{minipage}[b]{0.43\columnwidth}
        \includegraphics[width=0.98\columnwidth]{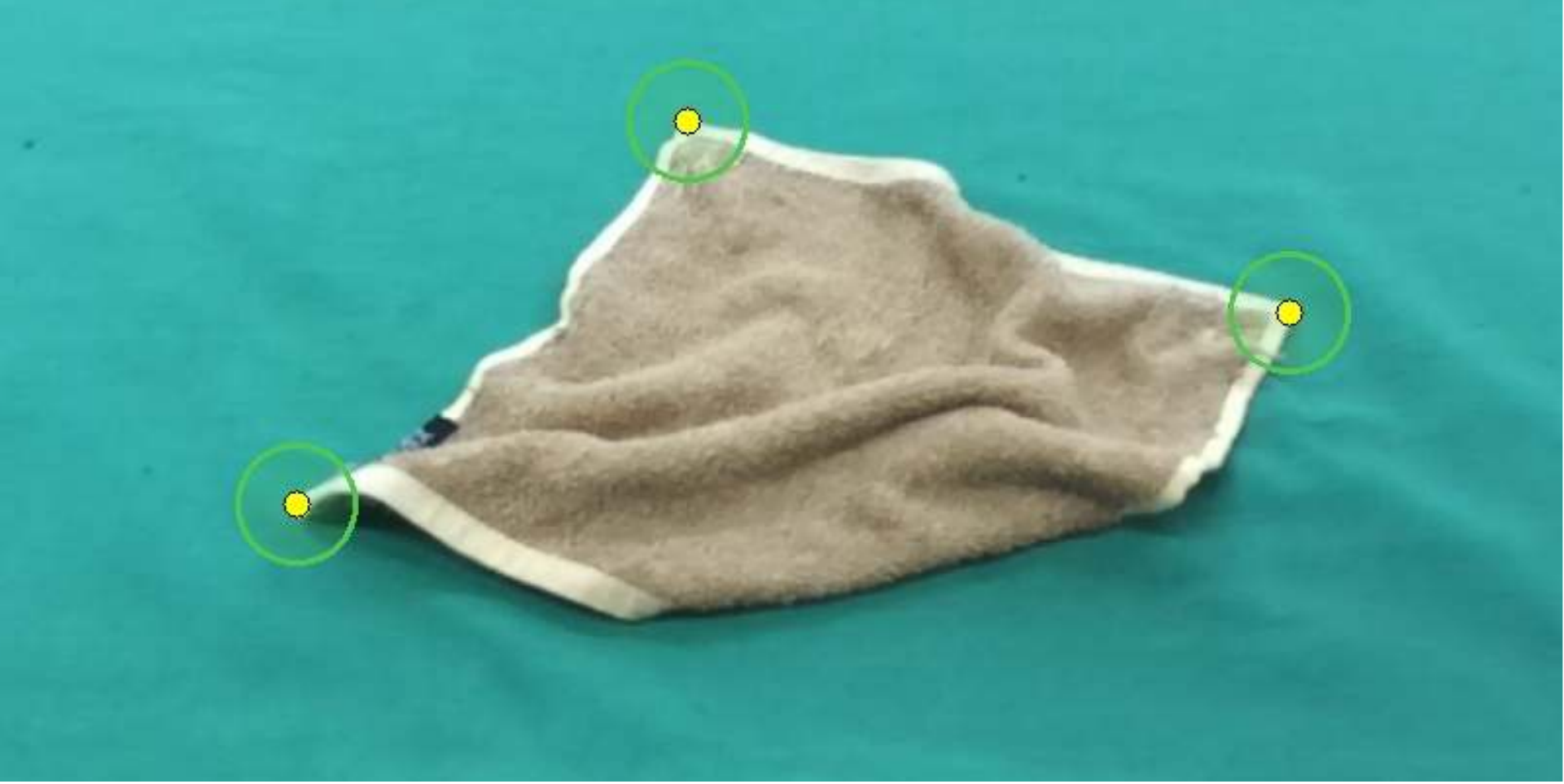}
    \end{minipage}
    \caption{Examples of observations with ground-truth criteria ($\epsilon = 30$ px)}
    \label{fig:example_recog_criteria}
    \vspace{-0.7em}
\end{figure}

\subsection{Ground-Truth Criteria for Valid Recognition}
\label{subsec:gt_criteria}
To evaluate the recognition efficacy of the integration of VLM-based ADP and Exploration Action, we established geometric ground-truth criteria for what constitutes valid recognition. For evaluation, we define valid recognition criteria as follows:
\begin{itemize}
    \item Each element $\mathbf{p}_i$ is within geometric tolerance $\epsilon$ of its ground-truth location $\mathbf{p}_i^{\text{gt}}$:
    \begin{equation}
        \|\mathbf{p}_i - \mathbf{p}_i^{\text{gt}}\| < \epsilon
    \end{equation}
\end{itemize}
The threshold $\epsilon$ defines the required recognition accuracy and is based on the grasping validity of the Preparation Action. 

In this study, we instantiate the valid recognition criteria as follows (Examples are shown in Fig.~\ref{fig:example_recog_criteria}):
\begin{itemize}
    \item \textit{Rope}: Both endpoints are represented within $\epsilon = 30$ px of their annotated ground-truth locations
    \item \textit{Cloth}: Two adjacent corners are represented within $\epsilon = 30$ px of their annotated ground-truth locations
\end{itemize}
where 30 px corresponds to approximately 2 cm in the real world.
These criteria are only for the evaluation of recognition validity. Our framework does not use these criteria during task execution.

\subsection{Metrics}
We defined evaluation metrics to assess (i) the validity of the integration of VLM-based ADP and Exploration Action (EA) for recognition efficacy, and (ii) the robustness of ExBot in task completion.

\subsubsection{Validity of ADP and EA}
To evaluate how the integration of the VLM-based ADP and Exploration Action 
resolves unrecognizability, we employed the following metrics:
\begin{itemize}
\item \textbf{Recognizability Rate (RR)}:  
This metric evaluated the effectiveness of the integration of ADP and 
Exploration Actions in transitioning OOD states into recognizable states $\mathcal{R}$ after $k$ exploration steps:
\begin{equation}
\text{RR}(k) = \frac{TP(k) + FN(k)}{N}
\end{equation}
where $TP$ (True Positive) denotes states that satisfy the criteria and correctly 
judged, $FN$ (False Negative) denotes states incorrectly judged as unrecognizable. $N$ is the total number of episodes. 

For episodes terminating before step $k$, the final recognition result (TP or FP) is carried forward.

\item \textbf{Cumulative Accuracy Rate (CAR)}:
This metric evaluates ADP's judgment accuracy:
\begin{equation}
\text{CAR}(k) = \frac{TP(k) + TN(k)}{N}
\end{equation}
where $TN$ (True Negative) denotes states that do not satisfy the ground-truth criteria and correctly judged.

For episodes terminating before step $k$, the final recognition result (TP or FP) is carried forward.

\item \textbf{False Negative Rate (FNR)}:
This auxiliary metric evaluates the conservative tendency of the ADP:
\begin{align}
\text{FNR}(k) = \frac{FN(k)}{TP(k) + FN(k)}
\end{align}
\end{itemize}

\subsubsection{Task Completion Robustness}
To evaluate the overall robustness, we measured success rates at the following sequential checkpoints:
\begin{itemize}
    \item \textbf{Transition Rate}: The percentage of trials where the system successfully transitioned the object to a bottleneck state $s_b$ via a Preparation Action $a_p$.
    \item \textbf{Task Completion Rate}: The percentage of trials where the system complete the manipulation task, given a successful transition to $s_b$.
    \item \textbf{Final Success Rate}: The overall end-to-end success rate across all OOD trials ($N=30$).
\end{itemize}

\subsection{Results and Evaluation}

\subsubsection{Validity of OOD State Creation Protocol}
Table~\ref{tab:orm_ability} shows the Object Recognition Module (ORM)'s performance on the same 50 OOD states. These results demonstrate that our protocol effectively generates severely self-occluded states that challenge standard perception systems, as evidenced by low initial recognition rates (less than 50\%).

\subsubsection{Importance of ORM-EA Complementarity for Robust Recognition}
To show that properly integrating ADP and Exploration Actions (EA) contributes to recognition improvement, we compare ExBot against the following baseline:
\begin{itemize}
    \item \textbf{VLM-based ORM w/ ADP and EA}: Uses the same ADP and Exploration Actions as ExBot, but employs a VLM-based object recognition module instead of our algorithm-based ORM. We used Qwen2.5-VL \cite{bai2025qwen25vltechnicalreport}, a VLM proficient in spatial pointing tasks.
\end{itemize}
This baseline shares ExBot's ADP and EA, isolating the impact of ORM design. Our EAs were designed to complement the explicit geometric assumptions of the algorithm-based ORM. By testing the same ADP and EA with a VLM-based ORM, whose internal recognition conditions are unclear, this comparison evaluated whether such design complementarity is essential for robust recognition performance, as shown in Table~\ref{tab:comparison}.

\textbf{Evaluation Framework:}
We evaluated system performance using two complementary metrics: Recognizability Rate (RR) and Cumulative Accuracy Rate (CAR). Their relationship characterizes system behavior:
\begin{itemize}
    \item \textit{High RR + High CAR}: The system successfully transitions objects into states compliant with the criteria and accurately recognizes keypoints
    \item \textit{High RR + Low CAR}: The system achieves transition, but ADP exhibits a high False Negative Rate (FNR), indicating conservative judgment
    \item \textit{Low RR + High CAR}: ADP identifies unrecognizable states, but Exploration Action fails to improve recognizability, indicating the poor ORM-EA performance
    \item \textit{Low RR + Low CAR}: Failure across all components
\end{itemize}

\begin{figure}
    \centering
    \begin{minipage}{0.98\linewidth}
        \includegraphics[width=0.98\linewidth]{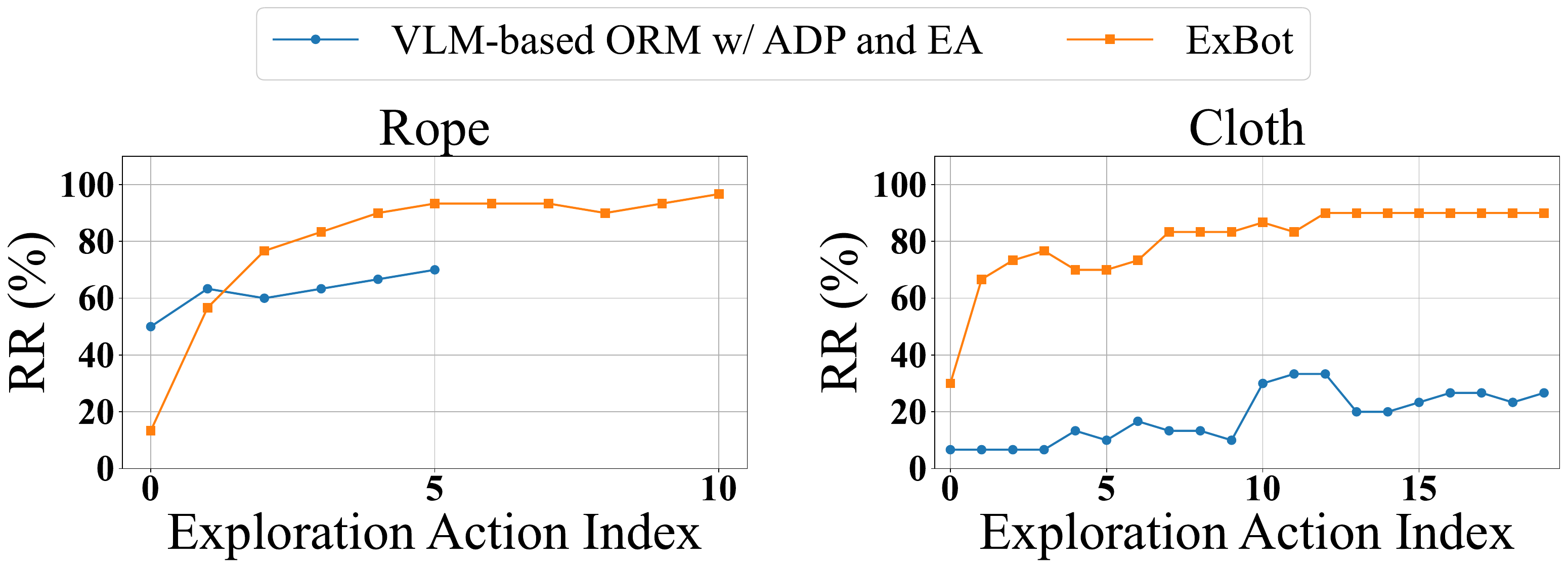}
    \end{minipage}
    \caption{Recognizability Rate (RR) as a function of the number of Exploration Actions for rope and cloth manipulation. This metric illustrates how the integration of ADP and EA enhances object recognizability for different ORMs.}
    \label{fig:recognizability_rate}
\end{figure}

\textbf{Recognizability Rate Analysis:}
Figure~\ref{fig:recognizability_rate} shows the Recognizability Rate (RR) as a function of exploration steps for both methods. RR(0) means the recognition success rate of the ORMs at the initial OOD states, and the improvement of RR reflects the influence of EA against different ORMs.

For rope manipulation, ExBot achieves a steady improvement of 84\%, reaching 97\% by step 10, whereas the VLM-based baseline exhibits a moderate success rate but plateaus at around 70\%.
For cloth manipulation, ExBot's RR increases dramatically from 30\% to over 80\% within 10 exploration steps, while the baseline remains below 35\% throughout.

\begin{figure}
    \centering
    \begin{minipage}{0.98\linewidth}
        \includegraphics[width=0.98\linewidth]{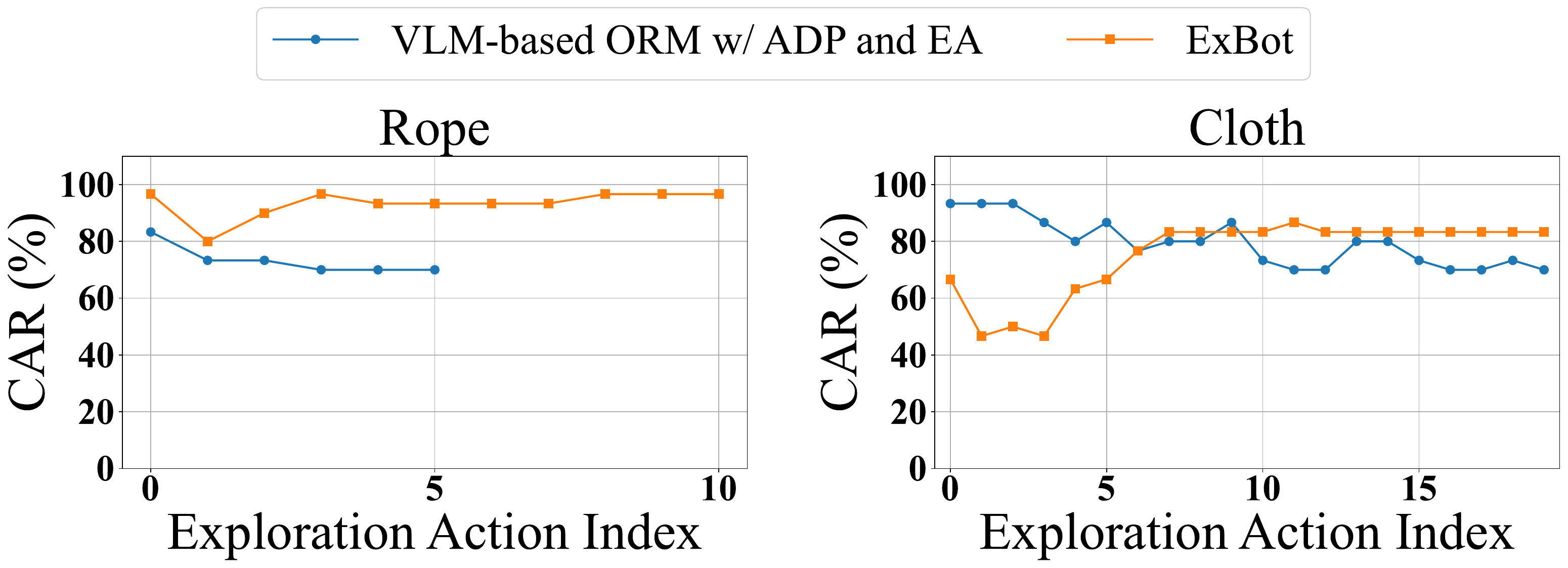}
    \end{minipage}
    \caption{Cumulative Accuracy Rate (CAR) as a function of the number of Exploration Actions for rope and cloth manipulation. This shows how ADP correctly infers recognizability throughout the recognition process.}
    \label{fig:cumulative_accuracy}

    \vspace{-0.7em}
\end{figure}

\textbf{Cumulative Accuracy Rate Analysis:}
Figure~\ref{fig:cumulative_accuracy} presents the Cumulative Accuracy Rate (CAR), measuring ADP's correctness in judging recognizability. Critical differences emerge as exploration progresses. The VLM-based baseline exhibits a high yet declining CAR trend (rope: 83\% $\rightarrow$ 70\%, cloth: 93\% $\rightarrow$ 70\%), mainly caused by accumulating False Positives as exploration steps progress. In contrast, ExBot reaches a stable or recovering high CAR (rope: stable at around 90\%, cloth: 47\% $\rightarrow$ 83\%).
For cloth manipulation in ExBot, the initial CAR drop reflects ExBot's tendency toward conservative recognition judgments, which shows high FNR during early exploration steps (0.65--0.75 for 1--3 steps).

\begin{figure}[t]
  \centering
  \begin{minipage}{0.95\linewidth}
    \centering
    \renewcommand{\arraystretch}{1.2}
    \captionof{table}{Object Recognition Module performance}
    \label{tab:orm_ability}
    \fontsize{7.5}{9}\selectfont
    \setlength{\tabcolsep}{3pt}
    \begin{tabular}{lcc}
      \toprule
      \textbf{Method} & \textbf{Rope} & \textbf{Cloth} \\ \midrule
      VLM-based ORM w/o ADP and EA & 22/50 & 3/50 \\
      ExBot w/o ADP and EA & 6/50 & 17/50 \\ \bottomrule
    \end{tabular}
  \end{minipage}

  \vspace{1em}

  \begin{minipage}{0.95\linewidth}
    \centering
    \renewcommand{\arraystretch}{1.2}
    \captionof{table}{Comparison of recognition accuracy from 30 OOD states}
    \label{tab:comparison}
    \fontsize{7.5}{9}\selectfont
    \setlength{\tabcolsep}{3pt}
    \begin{tabular}{lcccc}
      \toprule
      \textbf{Method} & \multicolumn{2}{c}{\textbf{Rope}} & \multicolumn{2}{c}{\textbf{Cloth}} \\ 
      \cmidrule(lr){2-3} \cmidrule(lr){4-5}
       & \textbf{Initial} & \textbf{Final} & \textbf{Initial} & \textbf{Final} \\ \midrule
      VLM-based ORM w/ ADP and EA & 12/30 & 21/30 & 0/30 & 6/30 \\
      \textbf{ExBot (Ours)} & 2/30 & \textbf{29/30} & 0/30 & \textbf{25/30} \\ \bottomrule
    \end{tabular}
  \end{minipage}

  \vspace{1em}

  \begin{minipage}{0.95\linewidth}
    \centering
    \renewcommand{\arraystretch}{1.2}
    \captionof{table}{Success rates at each stage ($N=30$)}
    \label{tab:success_rates}
    \fontsize{7.5}{9}\selectfont
    \setlength{\tabcolsep}{3pt}
    \begin{tabular}{lcccc}
      \toprule
      \textbf{Evaluation Phase}
        & \multicolumn{2}{c}{\textbf{End-to-End}}
        & \multicolumn{2}{c}{\textbf{ExBot (Ours)}} \\
      \cmidrule(lr){2-3} \cmidrule(lr){4-5}
        & \textbf{Rope} & \textbf{Cloth} & \textbf{Rope} & \textbf{Cloth} \\ \midrule
      Transition      & --- & --- & 23/30 & 23/30 \\
      Task Completion & 0/30  & 0/30  & \textbf{18/23} & \textbf{16/23} \\ \midrule
      \textbf{Final Success}
        & \textbf{0/30} & \textbf{0/30}
        & \textbf{18/30} & \textbf{16/30} \\ \bottomrule
    \end{tabular}
  \end{minipage}

    \vspace{-1em}
\end{figure}

These patterns are reflected in the final recognition outcomes (Table~\ref{tab:comparison}). ExBot achieves a final recognition accuracy (the rate of True Positives) of 29/30 for rope and 25/30 for cloth, outperforming the VLM-based baseline (21/30 and 6/30, respectively). Although the baseline exhibits initially high CAR, it shows limited RR improvement. Therefore, Exploration Actions fail to restore the VLM-based ORM's recognition conditions. Consequently, this failure leads to an increase in False Positives, which in turn causes the declining CAR trend. In contrast, ExBot systematically restores these conditions through ORM-EA complementarity, enabling substantial RR improvement while maintaining ADP accuracy. The algorithm-based ORM has explicit assumptions (e.g., most segments are visible), enabling the systematic design of complementary Exploration Actions. However, the VLM-based ORM lacks clear specifications of recognition conditions. Consequently, the baseline showed limited improvement (from 12/30 to 21/30 for rope, from 0/30 to 6/30 for cloth), while ExBot achieves substantial improvement (from 2/30 to 29/30 for rope, from 0/30 to 25/30 for cloth).

These results indicate that the ORM-EA complementarity enabled robust recognition. The co-designed approach prevented the ADP accuracy degradation observed in weakly-coupled systems such as the baseline, where exploration failed to generate views that reliably support recognition, leading to a declining CAR trend despite the initially high accuracy.

\subsubsection{Data Efficiency with Bottleneck Transition}
\label{subsubsec:bc_model}
As shown in Table~\ref{tab:success_rates}, the Transition Rate achieves 76.7\% (23/30) for both tasks, while the Task Completion Rate reaches 78.3\% (18/23) for rope and 69.6\% (16/23) for cloth with 157 and 138 demonstrations. These results validate the concept of bottleneck states transition and the downstream policy reliability from bottleneck states.

For comparison, we trained an end-to-end baseline learning the full trajectory from OOD initial states to task completion via bottleneck states, using identical demonstration data (Sec.~\ref{subsec:task_def_data_collect}). As shown in Table~\ref{tab:success_rates}, the end-to-end model failed to grasp object endpoints or corners, resulting in complete task failure (0/30 for both tasks). This demonstrates that end-to-end approaches cannot handle diverse OOD states with comparable training data. However, by constraining the initial state distribution to a standardized bottleneck, ExBot reduces deformation complexity and enables robust handling of deformable objects with significantly less data.

\subsection{Failure Analysis}
\label{subsec:failure_analysis}
Although our framework improved robustness against OOD states, we identified specific failure modes across the pipeline stages (summarized in Table~\ref{tab:failure_analysis}).

\textbf{Recognition Failures:}
Even with VLM-based ADP and Exploration Actions, the system failed to obtain an accurate representation before the Preparation Action (1/30 for rope, 5/30 for cloth). For the rope, the threshold $\epsilon$ classified one episode as a failure, subsequently causing the Preparation Action to fail. For the cloth, failures fall into three categories: First, the threshold $\epsilon$ led to preparation failure (1/5). Second, the VLM exhibited overly conservative judgment, selecting unnecessary Exploration Actions and failing to infer within 20 attempts (2/5). Third, incorrect inferences resulted from misleading structural features. When encountering cloth folded in half, the VLM incorrectly inferred that the corners of the shorter hem were the cloth corners (2/5).

\textbf{Transition and Task Execution Failures:}
Even after successful recognition, transitions occasionally failed (6/29 for rope, 2/25 for cloth). These failures stem from pinch grasp errors caused by calibration inaccuracies and gripper clearance. This can be mitigated by improved calibration and more sophisticated Preparation Actions. For rope manipulation, grasping endpoints requires precise rotation along the rope direction while avoiding adjacent parts, rendering the Preparation Action highly sensitive to gripper clearance.

Task Completion failure (5/23 for rope, 7/23 for cloth) was attributed to covariate shift. Even starting from bottleneck states, small deviations can compound into execution errors caused by twisting or friction with the table. Therefore, while the bottleneck concept effectively addresses initialization, further improvement is needed for robust execution. This is particularly true for long-horizon DOM involving difficult behaviors such as re-grasping.

\begin{table}[t]
\centering
\small
\setlength{\tabcolsep}{4pt} 
\caption{Summary of failure modes across different pipeline stages for rope and cloth manipulation tasks.}
\label{tab:failure_analysis}
\begin{tabular}{lccl}
\toprule
\textbf{Stage} & \textbf{Object} & \textbf{Rate} & \textbf{Primary Causes} \\ \midrule
\textbf{Recognition} & \textbf{Rope} & 1/30 & over threshold $\epsilon$ \\
 & \textbf{Cloth} & 5/30 & over threshold $\epsilon$ (1/5) \\
 & & & VLM over-conservation (2/5) \\ 
 & & & misleading inference (2/5) \\ \midrule
\textbf{Transition} & \textbf{Rope} & 6/29 & Pinch grasp errors \\
 & \textbf{Cloth} & 2/25 &  \\ \midrule
\textbf{Task Completion} & \textbf{Rope} & 5/23 & Covariate shift \\
 & \textbf{Cloth} & 7/23 & \\ \bottomrule
\end{tabular}
\vspace{-1em}
\end{table}

\section{CONCLUSIONS}

We presented ExBot, a novel framework for DOM that achieves data-efficient learning via bottleneck states and embraces imperfect perception to handle diverse initial states. 

The core of our approach lies in reconceptualizing the OOD problem. In contrast to prior methods that require accurate models or exhaustive data, we decompose the problem of infinite configurations into a manageable set of transitions by defining bottleneck states and partitioning the state space based on recognizability. This enabled data-efficient learning through standardized configurations. Moreover, Exploration Actions also enabled ExBot to actively handle severe self-occlusions. Real-world experiments on rope and cloth confirmed that ExBot robustly handles complex OOD scenarios through the co-designed Object Recognition Module and Exploration Action. Moreover, a VLM-based ADP effectively guides bottleneck state transitions, enabling data-efficient learning that succeeds across diverse OOD states where end-to-end approaches with comparable data entirely fail.

However, limitations remain. First, OOD coverage does not yet extend to complex scenarios, such as knotted ropes or multiple objects in the scene. Second, for recognition, the system remains sensitive to the VLM's conservative judgments and the design of domain-informed Exploration Actions. Finally, regarding task completion, while the concept of bottleneck states affects performance, the specific sensitivity to these design choices remains unexplored.

Future work will focus on achieving a higher success rate by establishing a more sophisticated or adaptable design principle of complementarity between the Object Recognition Module and Exploration Action. Moreover, we will develop principles for more refined bottleneck states toward robust BC models. Furthermore, we aim to evaluate the framework's applicability across a wider range of complex objects, such as T-shirts.

Despite these challenges, our results indicate that embracing imperfect perception and focusing on bottleneck transitions pave the way for practical, data-efficient, and robust autonomous manipulation of deformable objects.

\begin{normalsize}
    \bibliographystyle{bib/IEEEtranBST/IEEEtran}
    \bibliography{bib/list}
\end{normalsize}

\end{document}

%% file: preamble.tex
\usepackage{graphics} 
\usepackage{mathptmx} 
\usepackage{times} 
\usepackage{amsmath} 
\usepackage{amssymb}  
\usepackage[ruled,vlined]{algorithm2e}

\usepackage[most]{tcolorbox}
\usepackage{xcolor}
\usepackage{capt-of}
\usepackage{svg}

\definecolor{promptbg}{RGB}{248, 249, 250}
\definecolor{promptframe}{RGB}{200, 200, 200}
\definecolor{highlight}{RGB}{230, 240, 255}

\newtcolorbox{promptbox}{
    colback=white,          
    colframe=black!70,      
    boxrule=0.5pt,          
    arc=2pt,                
    left=7pt, right=7pt,  
    top=5pt, bottom=5pt,
    width=\linewidth,       
    enhanced                
}

\usepackage{booktabs}